\documentclass[sigconf]{acmart}

\AtBeginDocument{%
  }

\usepackage{amsmath}
\usepackage{amssymb}
\usepackage{amsfonts}
\usepackage{algorithmic}
\usepackage{graphicx}
\usepackage{textcomp}
\usepackage{xcolor}

\usepackage{indentfirst}
\usepackage{color}
\usepackage{enumitem}
\usepackage{multirow}
\usepackage{colortbl}
\usepackage{tabularray}
\usepackage{graphicx}
\usepackage{subfigure}
\usepackage{graphicx, nicefrac}
\usepackage{url}
\usepackage[flushleft]{threeparttable}
\usepackage{bigstrut}
\usepackage{ifthen}

\usepackage{makecell}
\usepackage{diagbox}
\usepackage{soul}
\usepackage{tcolorbox}
\usepackage{colortbl}

\usepackage{booktabs} %
\usepackage{caption}
\usepackage[ruled,linesnumbered]{algorithm2e}

\usepackage{amsmath,amsfonts,bm}

\def\eqref#1{equation~\ref{#1}}

\def\1{\bm{1}}

\def\vh{{\bm{h}}}

\def\vx{{\bm{x}}}

\def\mX{{\bm{X}}}

\DeclareMathAlphabet{\mathsfit}{\encodingdefault}{\sfdefault}{m}{sl}
\SetMathAlphabet{\mathsfit}{bold}{\encodingdefault}{\sfdefault}{bx}{n}

\newcommand{\eg}{{\it e.g.}}

\newcommand{\ie}{{\it i.e.}}

\setcopyright{acmlicensed}
\copyrightyear{2025}
\acmYear{2025}
\acmDOI{XXXXXXX.XXXXXXX}

\acmISBN{978-1-4503-XXXX-X/2018/06}

\begin{document} \begin{sloppypar}

\title{The Power of Architecture: Deep Dive into Transformer Architectures for Long-Term Time Series Forecasting}

\settopmatter{authorsperrow=4}

\author{Lefei Shen}
\affiliation{%
  \institution{Zhejiang University}
  \city{Hangzhou}
  \country{China}
}
\email{lefeishen@zju.edu.com}

\author{Mouxiang Chen}
\affiliation{
  \institution{Zhejiang University}
  \city{Hangzhou}
  \country{China}}
\email{chenmx@zju.edu.cn}

\author{Han Fu}
\affiliation{
  \institution{Zhejiang University}
  \city{Hangzhou}
  \country{China}}
\email{fuhan.fh@alibaba-inc.com}

\author{Xiaoxue Ren}
\affiliation{
  \institution{Zhejiang University}
  \city{Hangzhou}
  \country{China}}
\email{xxren@zju.edu.cn}

\author{Xiaoyun Joy Wang}
\affiliation{
  \institution{State Street Technology \\ (Zhejiang) Ltd.}
  \city{Hangzhou}
  \country{China}}
\email{xiaoyun99@gmail.com}

\author{Jianling Sun}
\affiliation{
  \institution{Zhejiang University}
  \city{Hangzhou}
  \country{China}}
\email{sunjl@zju.edu.cn}

\author{Zhuo Li}
\authornote{Corresponding authors.}
\affiliation{
  \institution{State Street Technology \\ (Zhejiang) Ltd.}
  \city{Hangzhou}
  \country{China}}
\email{lizhuo@zju.edu.cn}

\author{Chenghao Liu}
\authornotemark[1]
\affiliation{
  \institution{Salesforce Research Asia}
  \country{Singapore}}
\email{twinsken@gmail.com}

\renewcommand{\shortauthors}{Shen et al.}

\begin{abstract}
Transformer-based models have recently become dominant in Long-term Time Series Forecasting (LTSF), yet the variations in their architecture, such as encoder-only, encoder-decoder, and decoder-only designs, raise a crucial question: What Transformer architecture works best for LTSF tasks?
However, existing models are often tightly coupled with various time-series-specific designs, making it difficult to isolate the impact of the architecture itself. 
To address this, we propose a novel taxonomy that disentangles these designs, enabling clearer and more unified comparisons of Transformer architectures. 
Our taxonomy considers key aspects such as attention mechanisms, forecasting aggregations, forecasting paradigms, and normalization layers. 
Through extensive experiments, we uncover several key insights: bi-directional attention with joint-attention is most effective; more complete forecasting aggregation improves performance; and the direct-mapping paradigm outperforms autoregressive approaches. 
Furthermore, our combined model, utilizing optimal architectural choices, consistently outperforms several existing models, reinforcing the validity of our conclusions.
We hope these findings offer valuable guidance for future research on Transformer architectural designs in LTSF.
Our code is available at {\color{blue} \url{https://github.com/HALF111/TSF_architecture}}.
\end{abstract}



\maketitle

\section{Introduction}

\par In recent years, Transformer-based models have become dominant in long-term time series forecasting (LTSF) tasks \cite{Informer, Autoformer, FEDformer,  PatchTST, iTransformer, TimeXer, ARMA_Attention, Pyraformer, TFT, PDFormer, BasisFormer, SAMformer, Scaleformer, Quatformer}, demonstrating strong performance across various real-world applications \cite{TSF_Energy_1, TSF_Energy_2, TSF_Economics_1, TSF_Web_1, TSF_Web_2, TSF_Web_3, TSF_Weather_1, TSF_Weather_2, TSF_Finance_1}.
Notably, these Transformer-based LTSF models exhibit great diversity in their architectures.
For instance, Informer \cite{Informer}, Autoformer \cite{Autoformer}, FEDformer \cite{FEDformer}, ETSformer \cite{ETSformer}, and Crossformer \cite{Crossformer} adopt an encoder-decoder architecture, separating the encoding and decoding processes.
In contrast, PatchTST \cite{PatchTST}, iTransformer \cite{iTransformer}, TimeXer \cite{TimeXer}, and Fredformer \cite{Fredformer} utilize an encoder-only design, directly projecting the encoder embeddings into the forecasting window. Additionally, ARMA-Attention \cite{ARMA_Attention} employs a decoder-only architecture, applying uni-directional attention to capture temporal dependencies and autoregressively generate predictions.
This architectural diversity raises a critical question: \textbf{What Transformer architecture works best for LTSF tasks?}

\begin{figure*}[t]
    \centering
    \includegraphics[width=0.875\textwidth]{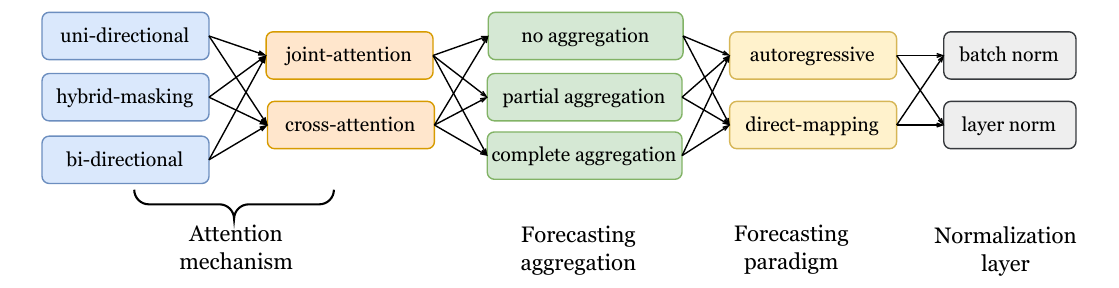}
    \vspace{-1.0em}
    \caption{Our proposed taxonomy on Transformer architectures for LTSF.}
    \label{fig:Transformer_macro_comparison}
    \vspace{-0.8em}
\end{figure*}

\par However, it is challenging to isolate the effect of the architecture itself, since many Transformer-based models are tightly coupled with various time-series-specific designs.
For example, Autoformer \cite{Autoformer} combines auto-correlation attention with seasonal-trend decomposition. 
FEDformer \cite{FEDformer} proposes Fourier frequency-enhanced attention to capture frequency-domain information. 
Crossformer \cite{Crossformer} incorporates two-stage attention to explicitly model both cross-time and cross-channel dependencies.
iTransformer \cite{iTransformer} inverts the attention mechanism and operates on the channel dimension to capture channel correlations.
Pathformer \cite{Pathformer} embeds adaptive multi-scale blocks to address multi-scale characteristics.
And Fredformer \cite{Fredformer} introduces frequency channel-wise attention for high-frequency feature extraction.
These coupled designs complicate direct comparisons of Transformer architectures, making it difficult to assess the architecture's independent impact on LTSF performance.

\par Despite the diversity of these time-series-specific designs, including seasonal-trend decomposition, frequency-domain information and channel correlation capturing, these models often blur the line between architecture itself and the auxiliary components.
To clarify the architectural distinctions, a unified taxonomy is needed to decouple them and isolate the influence of individual designs.

Furthermore, identifying the optimal Transformer architecture has also been a key focus in natural language processing (NLP), where the debate over encoder-only vs. decoder-only models \cite{LLM_arch_discussion_1, LLM_arch_discussion_2, LLM_arch_discussion_3, LLM_arch_discussion_4, LLM_arch_discussion_5} has evolved as decoder-only models dominate in large language models (LLMs), allowing researchers to focus on other interesting areas. Similarly, we believe that exploring the best architecture for LTSF can also guide future research and reduce experimental costs.

\begin{table*}[htbp]
  \centering
  \caption{The classification of existing Transformer-based LTSF models according to our taxonomy.}
  \vspace{-0.5em}
  \resizebox{0.92\linewidth}{!}{
    \begin{tabular}{c|p{10.085em}|c|c|c|c|c|c}
    \toprule
    \multicolumn{2}{c|}{\textbf{Models}} & \multicolumn{1}{c|}{\makecell{Vanilla- \\ Transformer \cite{Transformer}}} & \multicolumn{1}{p{8.165em}|}{\makecell{Informer \cite{Informer}, \\ Autoformer \cite{Autoformer}, \\ FEDformer \cite{FEDformer}, \\ Crossformer \cite{Crossformer}}} & \multicolumn{1}{p{7.915em}|}{\makecell{TST \cite{TST}, \\ PatchTST \cite{PatchTST}}} & \multicolumn{1}{c|}{\makecell{iTransformer \cite{iTransformer}, \\ Pathformer \cite{Pathformer}, \\ TimeXer \cite{TimeXer}, \\ TSLANet \cite{TSLANet}, \\ Fredformer \cite{Fredformer}}} & \multicolumn{1}{c|}{CATS \cite{CATS}} & \multicolumn{1}{c}{\makecell{ARMA- \\ Attention \cite{ARMA_Attention}}} \\
    \midrule
    \midrule
    \multirow{5}[4]{*}{\makecell{\textbf{Attention} \\ \textbf{mechanism}}} & \multicolumn{1}{c|}{bi-directional} &       &       & \multicolumn{1}{c|}{$\surd$} & \multicolumn{1}{c|}{$\surd$} & & \\
          & \multicolumn{1}{c|}{uni-directional} &       &       &       &   & \multicolumn{1}{c|}{$\surd$} & \multicolumn{1}{c}{$\surd$} \\
          & \multicolumn{1}{c|}{hybrid approach} & \multicolumn{1}{c|}{$\surd$} & \multicolumn{1}{c|}{$\surd$} &       &       &       &  \\
\cmidrule{2-8}          & \multicolumn{1}{c|}{joint-attention} &       &       & \multicolumn{1}{c|}{$\surd$} & \multicolumn{1}{c|}{$\surd$} & & \multicolumn{1}{c}{$\surd$} \\
          & \multicolumn{1}{c|}{cross-attention} & \multicolumn{1}{c|}{$\surd$} & \multicolumn{1}{c|}{$\surd$} & & & \multicolumn{1}{c|}{$\surd$} & \\
    \midrule
    \midrule
    \multirow{3}[2]{*}{\makecell{\textbf{Forecasting} \\ \textbf{aggregation}}} & \multicolumn{1}{c|}{no aggregation} & \multicolumn{1}{c|}{$\surd$} &       &       & & \multicolumn{1}{c|}{$\surd$} & \multicolumn{1}{c}{$\surd$} \\
          & \multicolumn{1}{c|}{partial aggregation} &       & \multicolumn{1}{c|}{$\surd$} & \multicolumn{1}{c|}{$\surd$} & \multicolumn{1}{c|}{$\surd$} & &  \\
          & \multicolumn{1}{c|}{complete aggregation} &       &       &       & & & \\
    \midrule
    \midrule
    \multirow{2}[2]{*}{\makecell{\textbf{Forecasting} \\ \textbf{paradigm}}} & \multicolumn{1}{c|}{auto-regressive} & \multicolumn{1}{c|}{$\surd$} &       &       & & & \multicolumn{1}{c}{$\surd$} \\
          & \multicolumn{1}{c|}{direct-mapping} &       & \multicolumn{1}{c|}{$\surd$} & \multicolumn{1}{c|}{$\surd$} & \multicolumn{1}{c|}{$\surd$} & \multicolumn{1}{c|}{$\surd$} &  \\
    \midrule
    \midrule
    \multirow{2}[2]{*}{\makecell{\textbf{Normalization} \\ \textbf{layer}}} & \multicolumn{1}{c|}{LayerNorm} & \multicolumn{1}{c|}{$\surd$} & \multicolumn{1}{c|}{$\surd$} &       & \multicolumn{1}{c|}{$\surd$} & \multicolumn{1}{c|}{$\surd$} & \multicolumn{1}{c}{$\surd$} \\
          & \multicolumn{1}{c|}{BatchNorm} &       &       & \multicolumn{1}{c|}{$\surd$} & & & \\
    \bottomrule
    \end{tabular}%
  }
  \label{tbl:existing_models_taxonomy}%
\end{table*}%

\par Meanwhile, although some previous studies have attempted to categorize and compare LTSF models, the question of the optimal Transformer architecture remains underexplored.
For instance, a survey on Transformers in time series forecasting \cite{Transformer_survey} reviews and classifies models from a methodological perspective, but lacks empirical analysis and does not offer conclusions or recommendations on optimal architectural designs.
Additionally, some benchmarking papers like BasicTS \cite{BasicTS}, TFB \cite{TFB}, and TSLib \cite{TimesNet} concentrate on aggregating time series datasets, establishing benchmarks, and empirically comparing existing models. However, they mainly provide high-level overviews, but fail to delve into the intrinsic architectural details. And they do not disentangle the influence of time-series-specific designs as well.

\par Therefore, given their insufficient experimental coverage and lack of in-depth architectural analysis, we propose a new and comprehensive taxonomy to investigate Transformer architectures for LTSF, accompanied by extensive empirical analysis to provide deeper insights.
Specifically, Figure \ref{fig:Transformer_macro_comparison} presents our taxonomy, and Table \ref{tbl:existing_models_taxonomy} classifies existing models according to it. Specifically, our taxonomy includes following dimensions:

\begin{itemize}[leftmargin=*]
    \item \textbf{(1) Attention mechanism.} 
    We first categorize attention mechanisms based on their masking strategy, including bi-directional non-causal attention \cite{PatchTST, iTransformer, TimeXer, Fredformer}, uni-directional causal attention \cite{ARMA_Attention, CATS}, and hybrid-masking approach \cite{Informer, Autoformer, FEDformer, Crossformer}. 
    We also distinguish between joint-attention (used in encoder-only or decoder-only models like \cite{PatchTST, iTransformer, Fredformer, ARMA_Attention}) and cross-attention (used in encoder-decoder models like \cite{Informer, Autoformer}). 
    Figure \ref{fig:01_Attention_mechanism} illustrates the differences between attention mechanisms.
    \item \textbf{(2) Forecasting aggregation.}
    Based on the extent of feature aggregation, we categorize forecasting aggregation approaches into: no aggregation \cite{Transformer, ARMA_Attention, CATS}, partial aggregation \cite{Informer, FEDformer, PatchTST, iTransformer}, and complete aggregation, as shown in Figure \ref{fig:02_feature_fusion}.
    
    \item \textbf{(3) Forecasting paradigm.} 
    We distinguish between the autoregressive paradigm \cite{ARMA_Attention} and the direct-mapping paradigm \cite{Informer, FEDformer, PatchTST} for LTSF, as illustrated in Figure \ref{fig:03_forecasting_objective}.
\end{itemize}

\par Additionally, our taxonomy also includes other variations in architectural elements such as normalization layers:
\begin{itemize}[leftmargin=*]
    \item \textbf{(4) Normalization layer.} 
    While LayerNorm is commonly used in vanilla Transformers \cite{Transformer}, LTSF tasks typically involve fixed input lengths, making both LayerNorm and BatchNorm viable. 
    Some models retain LayerNorm \cite{Informer, FEDformer, iTransformer}, while some use BatchNorm \cite{TST, PatchTST}. Detailed differences are discussed in Section \ref{subsubsec:norm_layer}.

\end{itemize}

\par Based on our taxonomy, we design experiments and evaluate them on long-term time series forecasting benchmarks, deriving the following key conclusions:
\begin{itemize}[leftmargin=*]
    \item \textbf{(1) Bi-directional attention with joint-attention is more effective.} (Related to Taxonomy 1). 
    Bi-directional attention can better capture temporal dependencies, and unified joint-attention outperforms separate cross-attention components, which both lead to better performance.
    
    \item \textbf{(2) More complete forecasting aggregation enhances performance.} (Related to Taxonomy 2). 
    Performance improves as models move from token-level forecasting with no aggregation to partial aggregation on forecasting tokens, and to complete aggregation of both look-back and forecasting windows.
    
    \item \textbf{(3) Direct-mapping significantly outperforms autoregressive forecasting paradigm.} (Related to Taxonomy 3). 
    Direct-mapping avoids error accumulation and inconsistencies between training and inference, greatly improving performance over the autoregressive approach.
\end{itemize}

\par Additionally, by combining the optimal designs from our conclusions, we construct a model that outperforms several existing models, including FEDformer \cite{FEDformer}, PatchTST \cite{PatchTST}, iTransformer \cite{iTransformer}, CATS \cite{CATS}, and ARMA-Attention \cite{ARMA_Attention}.
These models, despite their specific designs to extract time series characteristics, still exhibit limitations due to suboptimal architecture, which further validates our conclusions and the power of architectures.

\par Through our analysis, we aim to provide insights for future research on devising and selecting the best Transformer architecture for LTSF. Our contributions include:
\begin{itemize}[leftmargin=*]
    \item We propose a comprehensive taxonomy of Transformer architectures, and identify the most effective designs for LTSF through in-depth analyses.
    \item We examine Transformer-based LTSF models from multiple perspectives, including attention mechanisms, forecasting aggregation strategies, forecasting paradigms, and normalization layers.
    \item Several key conclusions are derived from our extensive experiments, which offer valuable insights for future research.
    \item Our combined model with optimal architectures consistently outperforms several existing models, further validating our conclusions and the power of architectures.
\end{itemize}

\section{Preliminaries}

\subsection{Time Series Forecasting}
\par We begin by defining multi-variate time series forecasting.
For a multi-variate time series with $M$ variables, let $\vx_{t} \in \mathbb{R}^M$ represent the value at $t$-th time step.
Then given a historical look-back window sequence $\mX_{t-L:t} = [\vx_{t-L}, \cdots, \vx_{t-1}] \in \mathbb{R}^{L\times M}$ with a window length of ~$L$, the forecasting task is to use $\mX_{t-L:t}$ to predict the future values in the forecasting window: $\hat{\mX}_{t:t+T} = [\hat\vx_{t}, \cdots, \hat\vx_{t+T-1}] \in \mathbb{R}^{T\times M}$, where ~$T$ is the forecasting window length. 
Time series forecasting models, denoted as $f$, are trained to optimize the mapping from the look-back window sequence to the forecasting window sequence, \ie ~ $\hat{\mX}_{t:t+T} = f(\mX_{t-L:t})$.

\subsection{Transformer Architecture for LTSF} \label{subsec:transformer_arch_intro}
\par The Transformer architecture \cite{Transformer}, originally designed for NLP tasks, has been widely adopted in LTSF domain. 
A typical Transformer consists of stacked blocks, each with multi-head self-attention, layer normalization, a two-layer feed-forward network, residual connections, and an additional cross-attention module in the decoder. 
For LTSF, Transformers process the look-back window values as input tokens and generate output tokens as predictions for the forecasting window. 
The detailed formulation of Transformer components is presented in Appendix \ref{sec:Transformer_formulation} due to page limit.

\begin{figure}[t]
    \centering
    \includegraphics[width=0.46\textwidth]{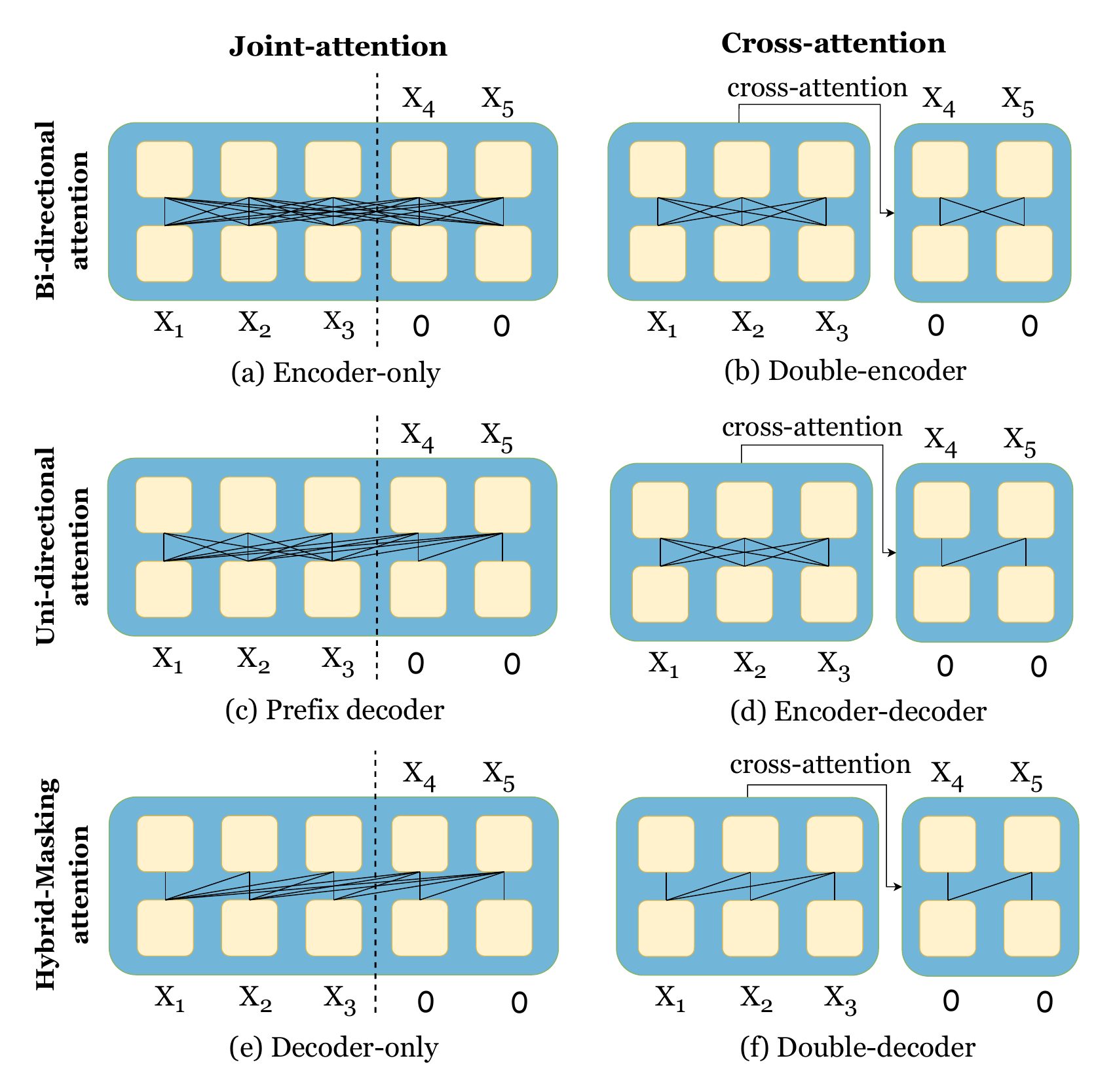}
    \vspace{-0.8em}
    \caption{
    Attention mechanism differences. Tokens $X_1 - X_3$ in the look-back window predict tokens $X_4 - X_5$ in the forecasting window. Positional embeddings are applied to all tokens but not explicitly shown, consistent across all figures.
    }
    \label{fig:01_Attention_mechanism}
    \vspace{-1.2em}
\end{figure}

\begin{figure}[t]
    \centering
    \includegraphics[width=0.46\textwidth]{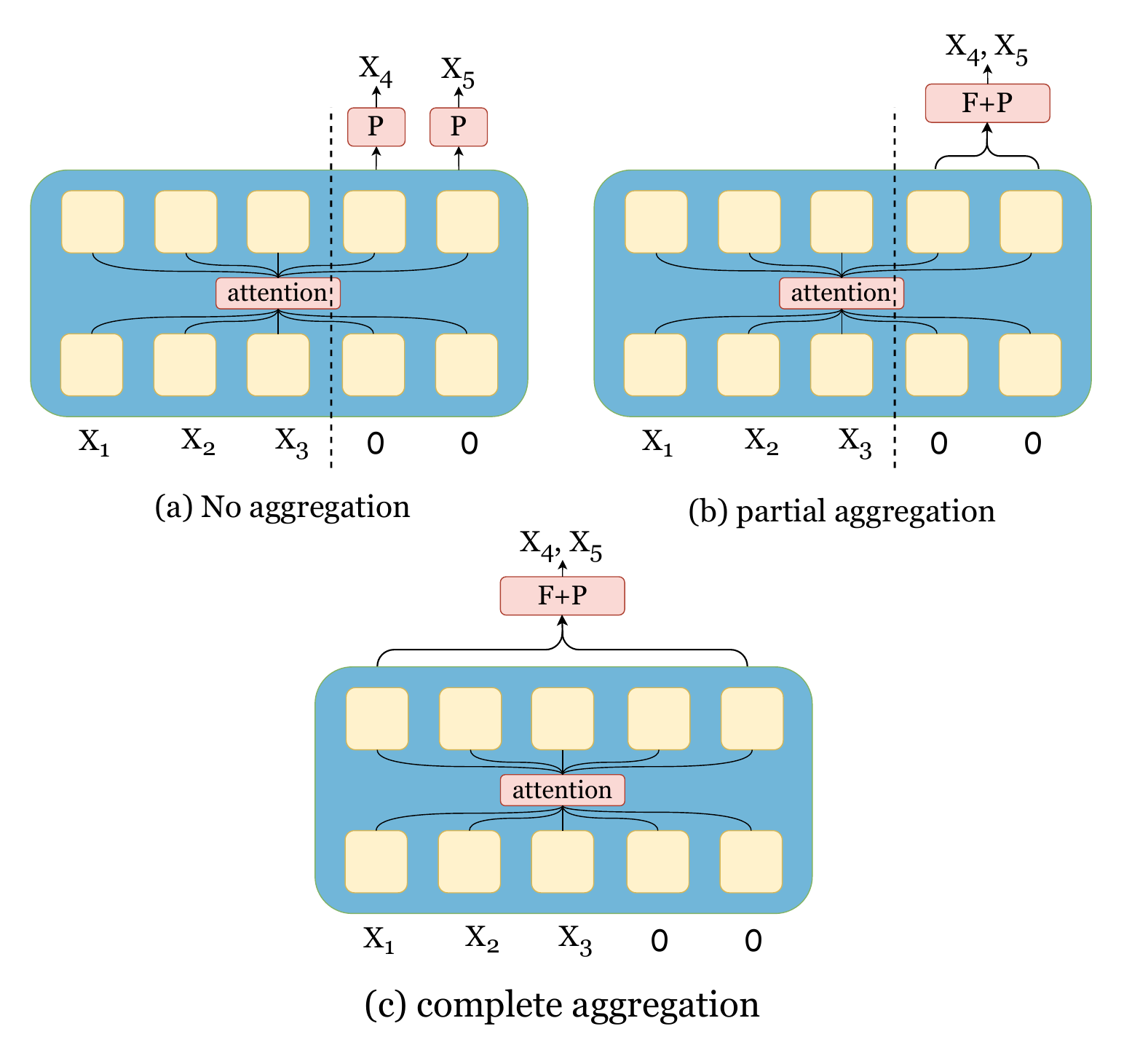}
    \vspace{-0.8em}
    \caption{
    Forecasting aggregation differences. In (a), ``\textit{P}'' represents a shared projection layer applied to each token. In (b)-(c), ``\textit{F+P}'' indicates ``\textit{Flatten+Projection}'', where token embeddings are flattened into a one-dimensional vector and then projected to the target values.
    }
    \label{fig:02_feature_fusion}
    \vspace{-1.2em}
\end{figure}

\begin{figure}[t]
    \centering
    \includegraphics[width=0.45\textwidth]{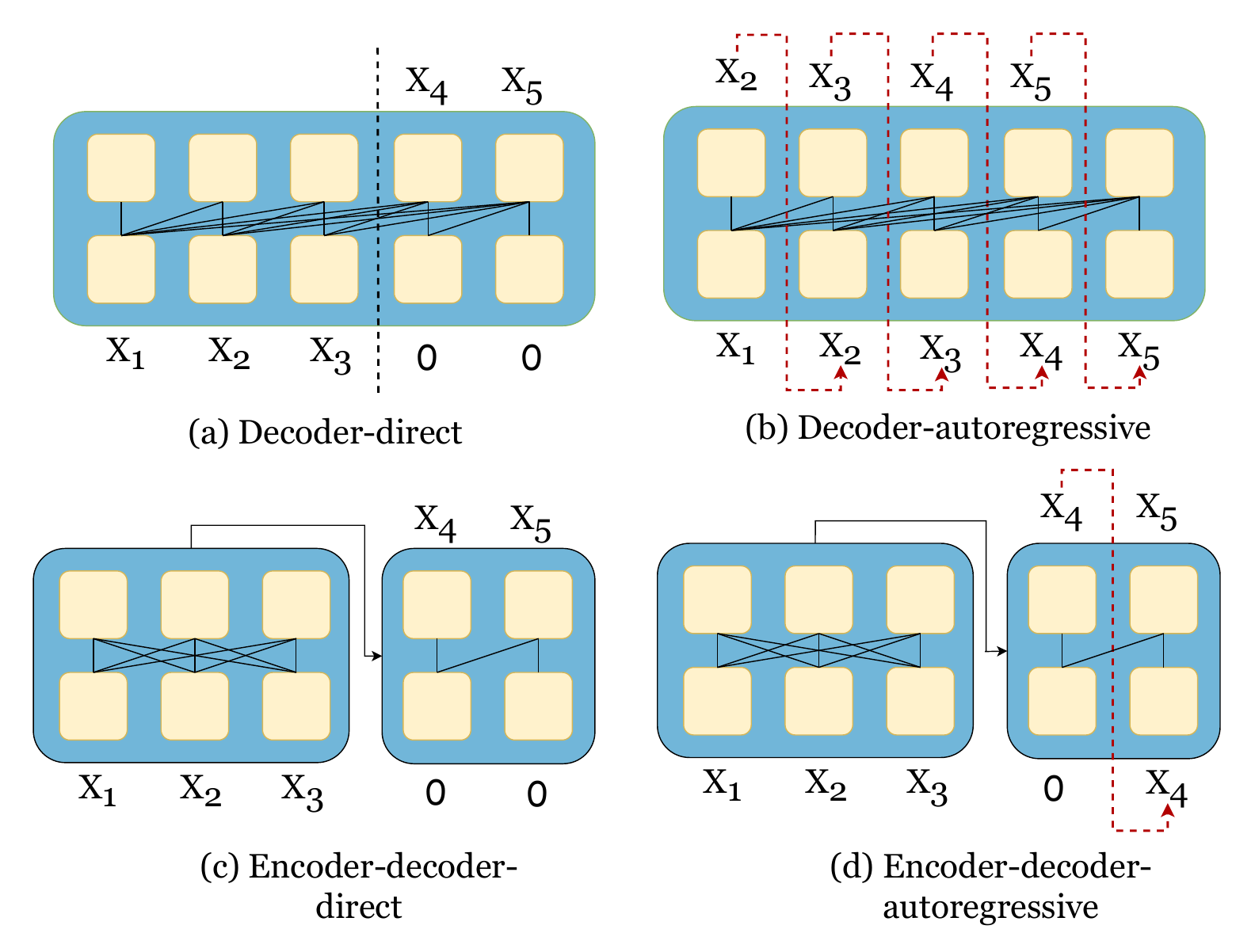}
    \vspace{-0.8em}
    \caption{
    Forecasting paradigm differences. Examples include decoder-only models with ``direct-mapping" (a), or with ``autoregressive'' (b) paradigms, and encoder-decoder models with similar patterns are in (c)-(d).
    }
    \label{fig:03_forecasting_objective}
    \vspace{-1.0em}
\end{figure}

\section{Architecture Designs}

\subsection{Taxonomy Protocols}

\subsubsection{Attention Mechanism} \label{subsubsec:attention_mechanism}
\
\par A key distinction among Transformer variants lies in the attention mechanism, which involves masking strategies applied to input tokens and the choice between joint-attention and cross-attention.

\par Masking strategies regulate the information flow between tokens, primarily including uni-directional causal masking, bi-directional non-causal masking, and hybrid-masking approach, which enables selective incorporation of contextual information.
\par The mathematical formula of attention score calculation on the distinction of masking strategy is as follows:
    \begin{itemize}[leftmargin=*]
        \item Bi-directional attention. Each token attends to all tokens.
        {\small
        \begin{equation}
        \mathrm{Attention}_i(\mathbf{q}_i, \mathbf{k}_j, \mathbf{v}_j) = \sum_{j=1}^L\alpha_{ij}\mathbf{v}_j, ~  \alpha_{ij} = \frac{\exp\left(\nicefrac{ \mathbf{q}_i\cdot\mathbf{k}_j}{\sqrt{d_k}}\right)}{\sum_{k=1}^L\exp\left(\nicefrac{\mathbf{q}_i\cdot\mathbf{k}_j}{\sqrt{d_k}} \right)}. \nonumber
        \end{equation}
        }
        where $L$ is sequence length, and $\mathbf{q}_i, \mathbf{k}_j, \mathbf{v}_j$ represent query, key, and value vectors, allowing each query $\mathbf{q}_i$ to attend to all keys $\{ \mathbf{k}_1, \ldots, \mathbf{k}_L \}$.
        
        \item Uni-directional attention. Each token is restricted to only attend to tokens at earlier positions: 
        {\small
        \begin{equation}
        \mathrm{Attention}_i(\mathbf{q}_i, \mathbf{k}_j, \mathbf{v}_j) = \sum_{j=1}^i\alpha_{ij}\mathbf{v}_j,~ \alpha_{ij}=\frac{\exp\left(\nicefrac{\mathbf{q}_i\cdot\mathbf{k}_j}{\sqrt{d_k}} \right)}{\sum_{k=1}^i\exp\left(\nicefrac{\mathbf{q}_i\cdot\mathbf{k}_j}{\sqrt{d_k}} \right)}. \nonumber
        \end{equation}
        }
        Here, the summation and softmax normalization are limited to $j \leq i$, so that each query $\mathbf{q}_i$ attends only to keys $\{ \mathbf{k}_1, \ldots, \mathbf{k}_i \}$.
        
        \item Hybrid-masking attention. This mechanism combines bi-directional and uni-directional attention by applying to different tokens in the look-back or the forecasting window.
    \end{itemize}

    \par Meanwhile, another critical distinction lies in the usage of joint-attention versus cross-attention:
    \begin{itemize}[leftmargin=*]
        \item Cross-attention. The model employs separate modules for tokens in look-back and forecasting windows, with the cross-attention module facilitating information interaction between them.
        \item Joint-attention. The model applies a unified self-attention module across tokens for both windows, enabling seamless interaction across the entire sequence.
    \end{itemize}

\par Combining these considerations, we categorize attention mechanisms into six distinct types, as illustrated in Figure \ref{fig:01_Attention_mechanism}. Each type is described in detail as follows.

\par \textbf{Encoder-only} (Figure \ref{fig:01_Attention_mechanism}(a)) employs bi-directional non-causal masking with joint-attention, allowing all tokens to interact across both look-back and forecasting windows. Output tokens are directly projected to the target values, enabling comprehensive feature extraction and interaction.

\par \textbf{Decoder-only} (Figure \ref{fig:01_Attention_mechanism}(e)) is widely used in LLMs like GPT \cite{GPT-2}, but less explored in LTSF. It employs uni-directional causal masking, limiting token interactions merely to past contexts. 
Output tokens are also directly projected to target values. We also discuss the autoregressive forecasting paradigm in Section \ref{subsubsec:forecasting_paradigm}.

\par \textbf{Prefix decoder} (Figure \ref{fig:01_Attention_mechanism}(c)) utilizes the hybrid-masking approach, which combines bi-directional attention for look-back window and uni-directional attention for forecasting window, offering richer representations than decoder-only models while moderately reducing computational costs compared to encoder-only models.

\par \textbf{Encoder-decoder} (Figure \ref{fig:01_Attention_mechanism}(d)) consists of two separate encoder and decoder components.
The encoder uses bi-directional self-attention on the look-back window, while the decoder employs uni-directional attention on the forecasting window, with cross-attention facilitating information transfer.

\par \textbf{Double-encoder \& Double-decoder} (Figure \ref{fig:01_Attention_mechanism}(b) \& \ref{fig:01_Attention_mechanism}(f)). 
These models modify the encoder-decoder model by substituting the attention mechanisms in either the encoder or decoder.
The double-encoder uses bi-directional attention in both modules, while the double-decoder uses uni-directional attention in both.

\subsubsection{Forecasting Aggregation} \label{subsubsec:forecasting_aggregation}
\
\par We then explore the forecasting aggregation approach, which is crucial in transforming latent representations into predictions.
A typical model has a feature extractor for extracting temporal features, and a prediction head to map these features to future values.
Recent studies \cite{PatchTST}, however, introduce feature interaction and aggregation within the prediction head, motivating us to explore different aggregation approaches. Figure \ref{fig:02_feature_fusion} illustrates three distinct types of forecasting aggregation.

\par \textbf{No aggregation} (Figure \ref{fig:02_feature_fusion}(a)). 
Each token’s latent embedding is independently mapped to target values via a shared projection layer, limiting token interaction and potentially hindering the modeling of complex dependencies. 
Formally, let $\hat{\mX}_{t:t+T} = [\hat\vx_{t}, \cdots, \hat\vx_{t+T-1}] \in \mathbb{R}^{T\times M}$ be the target values, and $f_{head}$ be the prediction head. For no aggregation, the latent embedding $\mathbf{H} = [\vh_{1}, \ldots, \vh_{T}]$ is in the shape of $\mathbb{R}^{T\times d_{model}}$, and the prediction head processes each time step independently: $\hat{\vx}_{t+i} = f_{head}(\vh_i),~ i \in \{1,\ldots,T\}$.

\par \textbf{Partial aggregation} (Figure \ref{fig:02_feature_fusion}(b)). 
Latent embeddings of forecasting window tokens are flattened and projected to target values, introducing token interaction within the forecasting window.
Formally, $\mathbf{H}$ is in the shape of $\mathbb{R}^{T\times d_{model}}$ or $\mathbb{R}^{L\times d_{model}}$ for partial aggregation, and the prediction head operates over the flattened embedding: $\hat{\mX}_{t:t+T} = f_{head}(\mathbf{H})$.

\par \textbf{Complete aggregation} (Figure \ref{fig:02_feature_fusion}(c)). 
Latent embeddings from both look-back and forecasting windows are flattened and projected to target values, enabling deeper interaction across the entire sequence and better capturing the long-range dependencies.
In this case, $\mathbf{H}$ spans both the look-back and forecasting windows, resulting in $\mathbb{R}^{(L+T)\times d_{model}}$, and the prediction head operates over the entire flattened embedding similarly: $\hat{\mX}_{t:t+T} = f_{head}(\mathbf{H})$.

\subsubsection{Forecasting Paradigms} \label{subsubsec:forecasting_paradigm}
\ 
\par Subsequently, the choice of forecasting paradigm also greatly affects model performance. 
Generally, there are two primary paradigms for generating values in the forecasting window: autoregressive and direct-mapping, as illustrated in Figure \ref{fig:03_forecasting_objective}.

\par \textbf{Autoregressive.} This paradigm predicts the next token iteratively based on previously generated tokens, as shown in Figures \ref{fig:03_forecasting_objective}(b) and \ref{fig:03_forecasting_objective}(d).
During training, the teacher-forcing technique \cite{teacher_forcing_1, teacher_forcing_2} uses the ground-truth as input at each step instead of predictions. 
While at inference, tokens are generated sequentially.
Formally, let the input sequence be $\mX_{t-L:t} = [\vx_{t-L}, \cdots, \vx_{t-1}]$ of length $L$, target sequence be $\hat{\mX}_{t:t+T} = [\hat\vx_{t}, \cdots, \hat\vx_{t+T-1}]$ of length $T$, and $f$ be the forecasting model. 
The autoregressive paradigm is defined as:
$$\hat\vx_{t} = f(\mX_{t-L:t}),$$
$$\hat\vx_{t+i} = f(\mX_{t-L:t}, \hat\vx_{t}, \ldots, \hat\vx_{t+i-1}), ~ i \in \{1,\ldots,T-1\}.$$

\par \textbf{Direct-mapping.} 
Unlike autoregressive, the direct-mapping paradigm generates all target tokens simultaneously in a single forward pass, avoiding the iterative process and reducing the error accumulation issue.
This paradigm is illustrated in Figures \ref{fig:03_forecasting_objective}(a) and \ref{fig:03_forecasting_objective}(c), where all target tokens ($X_4$ and $X_5$) are generated concurrently.
Formally, the direct-mapping paradigm is formulated as:
$$\hat{\mX}_{t:t+T} = f(\mX_{t-L:t}).$$

\subsubsection{Normalization Layer}
\label{subsubsec:norm_layer}
\ 
\par Normalization layers in Transformers ensure training stability and faster convergence. 
For LTSF, the choice between Layer Normalization (LN) and Batch Normalization (BN)
is essential due to their different characteristics and suitability for sequence modeling.

\par \textbf{Layer Normalization (LN).} 
LN normalizes across all features within each sample, making it suitable for variable-length input sequences. LTSF models like \cite{Informer, Autoformer, FEDformer} utilize LN, as in the original Transformer.

\par \textbf{Batch Normalization (BN).} 
BN normalizes across a batch of samples for each feature dimension, and leverages inter-sample information for regularization.
It is employed in LTSF models like TST \cite{TST} and PatchTST \cite{PatchTST}, where fixed input lengths in LTSF tasks enable meaningful batch statistics.

\par \textbf{Mathematical formulation.} 
Let the input to the normalization layer be $x \in \mathbb{R}^{B \times L \times d}$, where $B$ is the batch size, $L$ is the look-back window length, and $d$ is the latent embedding dimension. The output $y$ for both LN and BN can be expressed as:
$$y = \left( \frac{x - \mu(x)}{\sigma(x)} \right) \cdot \gamma + \beta,$$
where $\mu(x)$ and $\sigma(x)$ are the mean and standard deviation of $x$, and $\gamma$ and $\beta$ are learnable parameters.

\par The main distinction between LN and BN lies in the computation of $\mu(x)$ and $\sigma(x)$. For LN, they are computed across all features within each sample, resulting in $\mu(x)$ and $\sigma(x)$ with the shapes of $\mathbb R^{B}$.
For BN, they are computed across samples in a batch for each feature, yielding $\mu(x)$ and $\sigma(x)$ with the shapes of $\mathbb R^{d}$.

\subsection{Other Uniform Designs}

\par In this section, we introduce additional architectural designs consistently applied across all models for fair comparisons.

\begin{itemize}[leftmargin=*]
    \item \textbf{Standardization}: We employ z-score standardization for input processing, which is less sensitive to outliers \cite{Informer, FEDformer, PatchTST}.
    \item \textbf{RevIN}: Reversible Instance Normalization (RevIN) \cite{RevIn} normalizes each sample before prediction and then de-normalizes afterward \cite{PatchTST, ARMA_Attention, SAMformer}, addressing the distribution shift in time series.

    \item \textbf{Patching}: Following PatchTST \cite{PatchTST}, we utilize the patching approach, which aggregates consecutive timestamps into a single token to reduce token count and capture local dependencies \cite{PatchTST, TSLANet, Pathformer, TimeXer, ARMA_Attention}. Specifically, we employ a non-overlapping patching strategy since it better suits the autoregressive paradigm.

    \item \textbf{Channel handling strategy}: 
    For multi-variate time series, we apply the Channel Independence strategy, treating each channel as a separate uni-variate time series \cite{PatchTST, LTSF-Linear, ARMA_Attention, TSLANet, TimeXer, CATS}.

    \item \textbf{Positional embedding}:
    We use learnable additive positional embeddings \cite{PatchTST, ARMA_Attention, TSLANet, LogTrans} with the same dimensionality as token embeddings, to preserve temporal order across patches.
    
\end{itemize}

\section{Experiments and Analyses}

\begin{table*}[htbp]
  \centering
  \caption{Experiments on six attention mechanisms. All the results are averaged from four forecasting window length settings. Detailed results are presented in Table \ref{tbl:exp01_attention} in the appendix.}
  \vspace{-0.7em}
  \resizebox{0.95\linewidth}{!}{
    \begin{tabular}{c|cc|cc|cc|cc|cc|cc|cc|cc|c}
    \toprule
    \multicolumn{1}{c|}{\textbf{dataset}} & \multicolumn{2}{c|}{\textbf{Illness}} & \multicolumn{2}{c|}{\textbf{ETTh1}} & \multicolumn{2}{c|}{\textbf{ETTh2}} & \multicolumn{2}{c|}{\textbf{ETTm1}} & \multicolumn{2}{c|}{\textbf{ETTm2}} & \multicolumn{2}{c|}{\textbf{Weather}} & \multicolumn{2}{c|}{\textbf{ECL}} & \multicolumn{2}{c|}{\textbf{Traffic}} & \multirow{2}[4]{*}{\textbf{1st Count}} \\
\cmidrule{1-17}    \textbf{model} & \textbf{MSE} & \textbf{MAE} & \textbf{MSE} & \textbf{MAE} & \textbf{MSE} & \textbf{MAE} & \textbf{MSE} & \textbf{MAE} & \textbf{MSE} & \textbf{MAE} & \textbf{MSE} & \textbf{MAE} & \textbf{MSE} & \textbf{MAE} & \textbf{MSE} & \textbf{MAE} &  \\
    \midrule
    \midrule
    \textbf{Encoder-only} & \textbf{1.342 } & \textbf{0.766 } & \textbf{0.474 } & \textbf{0.448 } & 0.393  & 0.427  & \textbf{0.363 } & \textbf{0.383 } & 0.290  & 0.345  & 0.233  & \textbf{0.268 } & \textbf{0.158 } & \textbf{0.251 } & 0.385  & 0.263  & 9 \\
    \midrule
    \textbf{Prefix decoder} & 1.399  & 0.781  & 0.520  & 0.468  & 0.390  & 0.433  & 0.406  & 0.410  & 0.306  & 0.365  & 0.240  & 0.279  & \textbf{0.158 } & \textbf{0.251 } & \textbf{0.383 } & \textbf{0.261 } & 4 \\
    \midrule
    \textbf{Decoder-only} & 2.124  & 0.964  & 0.552  & 0.505  & 0.441  & 0.464  & 0.401  & 0.399  & 0.324  & 0.370  & 0.245  & 0.281  & 0.159  & 0.253  & 0.384  & 0.264  & 0 \\
    \midrule
    \textbf{Double-encoder} & 2.296  & 1.013  & 0.591  & 0.528  & \textbf{0.359 } & 0.418  & 0.391  & 0.400  & 0.343  & 0.382  & 0.310  & 0.343  & 0.187  & 0.278  & 0.404  & 0.300  & 1 \\
    \midrule
    \textbf{Encoder-decoder} & 1.835  & 0.816  & 0.510  & 0.463  & 0.361 & 0.406  & 0.402  & 0.405  & \textbf{0.287 } & \textbf{0.338 } & \textbf{0.229 } & 0.270  & 0.160  & 0.252  & 0.387  & 0.264  & 3 \\
    \midrule
    \textbf{Double-decoder} & 1.474  & 0.821  & 0.503  & 0.451  & 0.361  & \textbf{0.399 } & 0.417  & 0.414  & 0.289  & 0.341  & \textbf{0.229 } & 0.270  & 0.159  & 0.252  & 0.389  & 0.265  & 2 \\
    \bottomrule
    \end{tabular}%
    }
  \label{tbl:exp01_attention_smaller}%
\end{table*}%

\begin{figure*}[t!]
    \centering  %
    \vspace{-1.2em}
    \subfigure[Weather]{
        \includegraphics[width=0.275\textwidth]{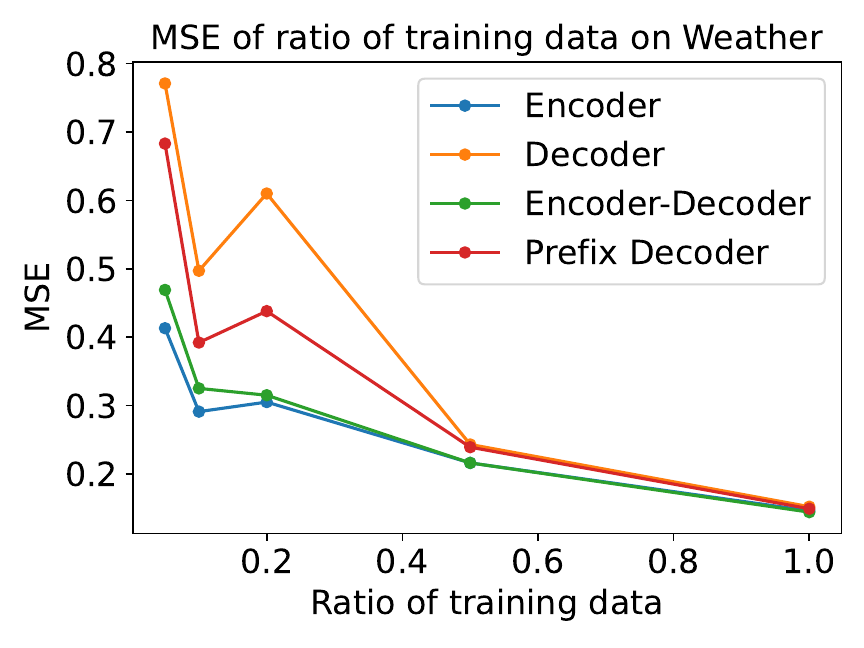}
    }%
    \subfigure[ECL]{
        \includegraphics[width=0.275\textwidth]{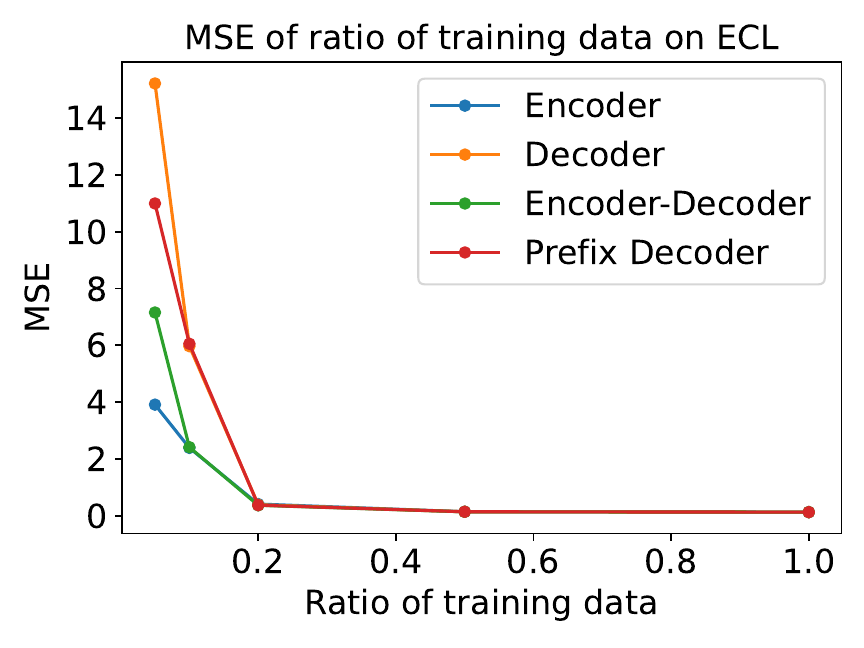}
    }%
    \subfigure[Traffic]{
        \includegraphics[width=0.275\textwidth]{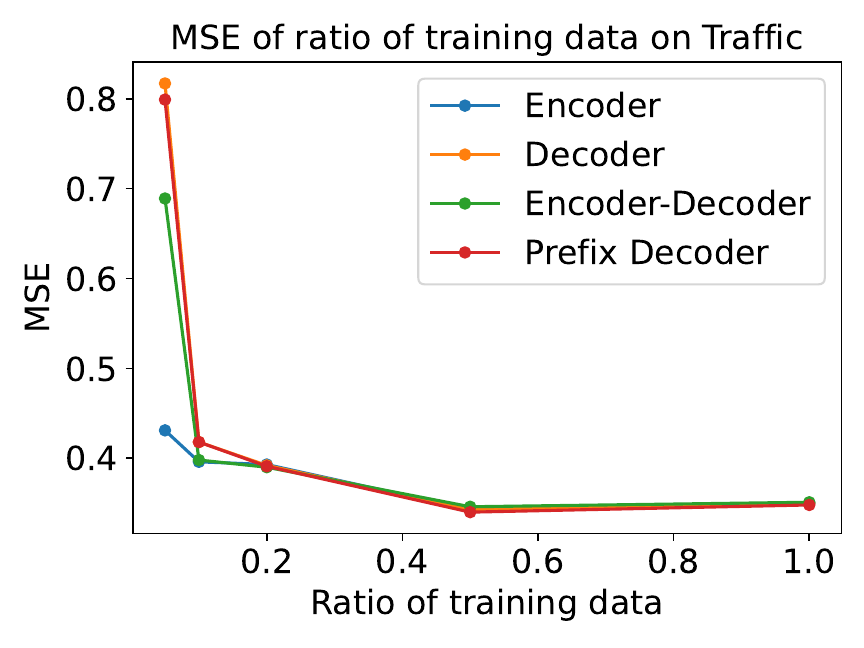}
    }%
    \vspace{-1.2em}
    \caption{MSE values on three datasets at different ratios of selected training samples to the original training samples.}
    \label{fig:train_ratio}
\end{figure*}

\subsection{Experiment Settings} \label{subsec:experiment_setting}
\par \textbf{Datasets.} We use eight widely used datasets for LTSF: Electricity, Traffic, Illness, Weather \cite{Autoformer}, and 4 ETT datasets (ETTh1, ETTh2, ETTm1, ETTm2) \cite{Informer}. Table \ref{tbl:dataset_statistics} in Appendix \ref{subsec:detailed_datasets} illustrates their statistics.
Following the common pre-processing protocols \cite{Informer, Autoformer}, we partition the datasets into train, validation, and test sets with a 6:2:2 ratio for 4 ETT datasets and 7:1:2 for the other datasets.

\par \textbf{Hyperparameters.} Key hyperparameters include forecasting window length $T$ of \{24,36,48,60\} for Illness and \{96,192,336,720\} for other datasets, look-back window length $L$ of 120 for Illness and 512 for others, non-overlapping patch length ($patch\_len$) of 6 for Illness and 16 for others, and a learning rate ($lr$) of 0.001.
Also, to guarantee consistent parameter counts, models with cross-attention (\eg, encoder-decoder, double-encoder, double-decoder) use 3+3 layers, while models with joint-attention use 6 stacked encoder or decoder layers.
Detailed settings are in Table \ref{tbl:main_parameters} in Appendix \ref{subsec:detailed_hyperparams}.

\par \textbf{Evaluation metrics.} We evaluate model performance using Mean Squared Error (MSE) and Mean Absolute Error (MAE), with lower values indicating better results.

\par \textbf{Forecasting Length Setting.} 
Most prior works use a fixed forecasting length setting, and train models to predict a specific output length.
However, inspired by NLP tasks like text generation where output length can vary, recent studies \cite{Moirai, GPT4TS, TimesFM, Timer} are exploring uniform forecasters for LTSF, capable of handling varying forecasting lengths.
Building on this idea, we propose a \textbf{variable forecasting length setting}, requiring models to predict multiple lengths, which is more challenging.
Autoregressive models handle this naturally due to their iterative nature, while direct-mapping models require training on the maximum length and selecting the initial portion of the output for shorter windows.

\subsection{Experimental Results} \label{subsec:experimental_results}
\par In this section, we present experiments based on each perspective of our taxonomy. For fair comparisons, when evaluating a specific architectural design such as the attention mechanism, all other design choices including forecasting aggregation, forecasting paradigm, and normalization layer, are kept consistent.

\subsubsection{Attention Mechanism} \label{subsubsec:exp_attention_mechanism}
\
\par We begin by comparing the attention mechanisms of six models in Figure \ref{fig:01_Attention_mechanism},
with results reported in Table \ref{tbl:exp01_attention_smaller}.

\par First, we analyze the masking strategies.
Performance improves progressively from decoder to prefix decoder to encoder models, reflecting a transition from uni-directional to hybrid-masking to bi-directional attention.
\textbf{This highlights that for masking strategies, bi-directional attention proves most effective.}
A reasonable explanation is that bi-directional attention allows each token to attend to both preceding and succeeding tokens, thus capturing temporal dependencies more effectively than uni-directional attention, which restricts token interaction.

\par Afterward, we examine the joint-attention and cross-attention.
Specifically, models with joint-attention (\eg, encoder, prefix decoder) outperform those with cross-attention (\eg, double-encoder, encoder-decoder) in most cases.
\textbf{This suggests that joint-attention is more effective than cross-attention.}
This is because joint-attention enables direct token-to-token connections, facilitating better historical context usage. While cross-attention limits token access to the look-back window via a single latent embedding, which may capture dependencies inadequately.

\par We also observe a large performance gap between encoder and double-encoder models, while a smaller difference between decoder and double-decoder models. This is likely due to uni-directional attention's limitations in modeling temporal dependencies, leading to poorer performance for both decoder and double-decoder models, thereby weakening the impact of cross-attention.

\begin{tcolorbox}[title = {Conclusion 1}] %
Bi-directional attention outperforms uni-directional attention, and joint-attention is superior to cross-attention. %
\end{tcolorbox}

\begin{table*}[htbp]
  \centering
  \caption{Experiments on three forecasting aggregation methods. All the results are averaged from four forecasting window length settings. Detailed results are presented in Table \ref{tbl:exp02_feature_fusion} in the appendix. }
  \vspace{-0.7em}
    \resizebox{0.95\linewidth}{!}{
    \begin{tabular}{c|cc|cc|cc|cc|cc|cc|cc|cc|c}
    \toprule
    \multicolumn{1}{c|}{\textbf{dataset}} & \multicolumn{2}{c|}{\textbf{Illness}} & \multicolumn{2}{c|}{\textbf{ETTh1}} & \multicolumn{2}{c|}{\textbf{ETTh2}} & \multicolumn{2}{c|}{\textbf{ETTm1}} & \multicolumn{2}{c|}{\textbf{ETTm2}} & \multicolumn{2}{c|}{\textbf{Weather}} & \multicolumn{2}{c|}{\textbf{ECL}} & \multicolumn{2}{c|}{\textbf{Traffic}} & \multirow{2}[4]{*}{\textbf{1st Count}} \\
\cmidrule{1-17}    \textbf{model} & \textbf{MSE} & \textbf{MAE} & \textbf{MSE} & \textbf{MAE} & \textbf{MSE} & \textbf{MAE} & \textbf{MSE} & \textbf{MAE} & \textbf{MSE} & \textbf{MAE} & \textbf{MSE} & \textbf{MAE} & \textbf{MSE} & \textbf{MAE} & \textbf{MSE} & \textbf{MAE} &  \\
    \midrule
    \midrule
    \textbf{No aggregation} 
        & 1.367  & 0.776  & 0.474  & 0.450  & 0.393  & 0.427  & 0.363  & 0.383  & 0.290  & 0.345  & 0.233  & 0.268  & 0.159  & 0.251  & 0.385  & 0.263  & 0 \\
    \midrule
    \textbf{Partial aggregation} 
        & 1.339  & 0.766  & 0.439  & 0.443  & 0.350  & 0.392  & 0.355  & 0.386  & 0.270  & 0.334  & 0.238  & 0.277  & 0.158  & 0.251  & 0.387  & 0.262  & 0 \\
    \midrule
    \textbf{Complete aggregation} 
        & \textbf{1.278 } & \textbf{0.740 } & \textbf{0.410 } & \textbf{0.433 } & \textbf{0.335 } & \textbf{0.377 } & \textbf{0.351 } & \textbf{0.374 } & \textbf{0.260 } & \textbf{0.324 } & \textbf{0.227 } & \textbf{0.265 } & \textbf{0.157 } & \textbf{0.250 } & \textbf{0.383 } & \textbf{0.259 } & 10 \\
    \bottomrule
    \end{tabular}%
    }
  \label{tbl:exp02_feature_fusion_smaller}%
\end{table*}%

\begin{table*}[htbp]
  \centering
  \caption{Experiments on forecasting paradigms for decoder-only and encoder-decoder models. All the results are averaged from four forecasting window length settings. Detailed results are presented in Table \ref{tbl:exp03_forecast_objective} in the appendix. }
  \vspace{-0.7em}
    \resizebox{0.96\linewidth}{!}{
    \begin{tabular}{c|cc|cc|cc|cc|cc|cc|cc|cc}
    \toprule
    \multicolumn{1}{c|}{\textbf{dataset}} & \multicolumn{2}{c|}{\textbf{Illness}} & \multicolumn{2}{c|}{\textbf{ETTh1}} & \multicolumn{2}{c|}{\textbf{ETTh2}} & \multicolumn{2}{c|}{\textbf{ETTm1}} & \multicolumn{2}{c|}{\textbf{ETTm2}} & \multicolumn{2}{c|}{\textbf{Weather}} & \multicolumn{2}{c|}{\textbf{ECL}} & \multicolumn{2}{c}{\textbf{Traffic}} \\
    \midrule
    \textbf{model} & \textbf{MSE} & \textbf{MAE} & \textbf{MSE} & \textbf{MAE} & \textbf{MSE} & \textbf{MAE} & \textbf{MSE} & \textbf{MAE} & \textbf{MSE} & \textbf{MAE} & \textbf{MSE} & \textbf{MAE} & \textbf{MSE} & \textbf{MAE} & \textbf{MSE} & \textbf{MAE} \\
    \midrule
    \midrule
    \textbf{Decoder-direct} 
        & \textbf{2.124 } & \textbf{0.964 } & \textbf{0.552 } & \textbf{0.505 } & \textbf{0.441 } & \textbf{0.464 } & \textbf{0.401 } & \textbf{0.399 } & \textbf{0.324 } & \textbf{0.370 } & \textbf{0.245 } & \textbf{0.281 } & \textbf{0.159 } & \textbf{0.253 } & \textbf{0.384 } & \textbf{0.264 } \\
    \midrule
    \textbf{Decoder-autoregressive} 
        & 5.318  & 1.652  & 1.233  & 0.766  & 0.622  & 0.540  & 1.352  & 0.749  & 0.598  & 0.511  & 0.543  & 0.461  & 0.465  & 0.410  & 5.318  & 1.652  \\
    \midrule
    \midrule
    \textbf{Encoder-decoder-direct} 
        & \textbf{1.835 } & \textbf{0.821 } & \textbf{0.510 } & \textbf{0.463 } & \textbf{0.359 } & \textbf{0.405 } & \textbf{0.402 } & \textbf{0.405 } & \textbf{0.287 } & \textbf{0.338 } & \textbf{0.229 } & \textbf{0.270 } & \textbf{0.160 } & \textbf{0.252 } & \textbf{0.387 } & \textbf{0.264 } \\
    \midrule
    \textbf{Encoder-decoder-autoregressive} 
        & 4.394  & 1.535  & 1.319  & 0.722  & 0.487  & 0.494  & 1.128  & 0.665  & 0.549  & 0.480  & 0.525  & 0.444  & 0.439  & 0.400  & 4.394  & 1.535  \\
    \bottomrule
    \end{tabular}%
    }
  \label{tbl:exp03_forecast_objective_smaller}%
\end{table*}%

\par Furthermore, we notice that although bi-directional attention consistently outperforms uni-directional attention, their performance gap varies across datasets. The gap is significant for Illness and four ETT datasets, but minimal for Weather, ECL, and Traffic.
Upon reviewing Table \ref{tbl:dataset_statistics} in \ref{subsec:detailed_datasets}, the latter three datasets have tens or hundreds of variates, generating more training samples due to the channel independence strategy. This suggests that the quantity of training samples also influences the performance gap.
\par To investigate this, we conduct additional experiments on Weather, ECL, and Traffic datasets by only varying the training data size.
Specifically, models are retrained using subsets from the most recent proportion of training data with ratios $\{0.05, 0.1, 0.2, 0.5, 1.0\}$, while the evaluation remains unchanged.
Results presented in Figure \ref{fig:train_ratio} lead to the following corollary: 

\begin{tcolorbox}[title = {Corollary 1}]
When data is limited, encoder-only models greatly outperform decoder-only models. However, as data size increases, their performance gap gradually diminishes.
\end{tcolorbox}

\par A plausible explanation is that uni-directional attention in decoder models restricts each token to attend only to preceding tokens, making the task inherently more challenging, especially with limited data.
However, with sufficient data, even uni-directional attention can fully learn the patterns and temporal dependencies, thus reducing the impact of attention mechanism differences.

\subsubsection{Forecasting Aggregation} \label{subsubsec:exp_forecasting_aggregation}
\
\par Afterward, we compare forecasting aggregation methods across three architectures in Figure \ref{fig:02_feature_fusion}, with results reported in Table \ref{tbl:exp02_feature_fusion_smaller}.

\par The results indicate that complete aggregation (``F+P" on all tokens) outperforms partial aggregation (``F+P" on forecasting tokens), which in turn outperforms no aggregation (Shared Projection). Based on this, we obtain the following conclusion:

\begin{tcolorbox}[title = {Conclusion 2}]
More complete and comprehensive feature aggregation across tokens in both look-back and forecasting windows enhances LTSF performance.
\end{tcolorbox}

\par A plausible explanation is that: in the attention module, token interaction occurs at the token-wise level. However, a complete aggregation in the prediction head flattens token embedding into a one-dimensional vector before projection, enabling interaction at both token-wise and feature-wise levels. This facilitates deeper feature aggregation and fusion, leading to improved performance.

\begin{table*}[htbp]
  \centering
  \caption{Experiments on the normalization layer, including LayerNorm (LN) and BatchNorm (BN). All the results are averaged from four forecasting window length settings. Detailed results are presented in Table \ref{tbl:exp04_norm_layer} in the appendix. }
  \vspace{-0.7em}
    \resizebox{0.92\linewidth}{!}{
    \begin{tabular}{c|c|cc|cc|cc|cc|cc|cc|cc|cc}
    \toprule
    \multicolumn{2}{c|}{\textbf{dataset}} & \multicolumn{2}{c|}{\textbf{Illness}} & \multicolumn{2}{c|}{\textbf{ETTh1}} & \multicolumn{2}{c|}{\textbf{ETTh2}} & \multicolumn{2}{c|}{\textbf{ETTm1}} & \multicolumn{2}{c|}{\textbf{ETTm2}} & \multicolumn{2}{c|}{\textbf{Weather}} & \multicolumn{2}{c|}{\textbf{ECL}} & \multicolumn{2}{c}{\textbf{Traffic}} \\
    \midrule
    \multicolumn{2}{c|}{\textbf{anomaly sample ratio}} & \multicolumn{2}{c|}{\textbf{0.322}} & \multicolumn{2}{c|}{\textbf{0.127}} & \multicolumn{2}{c|}{\textbf{0.282}} & \multicolumn{2}{c|}{\textbf{0.099}} & \multicolumn{2}{c|}{\textbf{0.160}} & \multicolumn{2}{c|}{\textbf{0.141}} & \multicolumn{2}{c|}{\textbf{0.034}} & \multicolumn{2}{c}{\textbf{0.025}} \\
    \midrule
    \textbf{model} & \textbf{norm} & \textbf{MSE} & \textbf{MAE} & \textbf{MSE} & \textbf{MAE} & \textbf{MSE} & \textbf{MAE} & \textbf{MSE} & \textbf{MAE} & \textbf{MSE} & \textbf{MAE} & \textbf{MSE} & \textbf{MAE} & \textbf{MSE} & \textbf{MAE} & \textbf{MSE} & \textbf{MAE} \\
    \midrule
    \midrule
    \multirow{2}[4]{*}{\textbf{Encoder-only}} & \multirow{1}[2]{*}{\textbf{+LN}} 
          & 1.429  & 0.830  & \textbf{0.408 } & \textbf{0.431 } & 0.346  & 0.385  & \textbf{0.347 } & \textbf{0.371 } & 0.269  & 0.328  & \textbf{0.227 } & \textbf{0.265 } & 0.158  & 0.252  & \textbf{0.382 } & \textbf{0.258 } \\
\cmidrule{2-18}          & \multirow{1}[2]{*}{\textbf{+BN}}
          & \textbf{1.278 } & \textbf{0.740 } & 0.410  & 0.433  & \textbf{0.335 } & \textbf{0.377 } & 0.351  & 0.374  & \textbf{0.260 } & \textbf{0.324 } & \textbf{0.227 } & \textbf{0.265 } & \textbf{0.157 } & \textbf{0.250 } & 0.383  & 0.259  \\
    \midrule
    \midrule
    \multirow{2}[2]{*}{\textbf{Encoder-decoder}} & \multirow{1}[1]{*}{\textbf{+LN}}
          & 1.440  & 0.793  & \textbf{0.510 } & \textbf{0.461 } & \textbf{0.357 } & 0.409  & \textbf{0.396 } & 0.406  & 0.294  & 0.342  & 0.233  & 0.274  & \textbf{0.159 } & \textbf{0.250 } & \textbf{0.386 } & \textbf{0.262 } \\
\cmidrule{2-18}          & \multirow{1}[1]{*}{\textbf{+BN}} 
          & \textbf{1.333 } & \textbf{0.764 } & \textbf{0.510 } & 0.463  & 0.359  & \textbf{0.406 } & 0.402  & \textbf{0.405 } & \textbf{0.289 } & \textbf{0.341 } & \textbf{0.229 } & \textbf{0.270 } & 0.160  & 0.252  & 0.387  & 0.264  \\
    \midrule
    \midrule
    \multirow{2}[3]{*}{\textbf{Decoder-only}} & \multirow{1}[1]{*}{\textbf{+LN}}
          & \textbf{1.770 } & \textbf{0.903 } & 0.553  & \textbf{0.504 } & 0.476  & 0.477  & \textbf{0.391 } & \textbf{0.397 } & 0.336  & 0.378  & 0.255  & 0.289  & 0.160  & 0.254  & 0.385  & \textbf{0.264 } \\
\cmidrule{2-18}          & \multirow{1}[2]{*}{\textbf{+BN}}
          & 2.124  & 0.964  & \textbf{0.552 } & 0.505  & \textbf{0.441 } & \textbf{0.464 } & 0.401  & 0.399  & \textbf{0.324 } & \textbf{0.370 } & \textbf{0.245 } & \textbf{0.281 } & \textbf{0.159 } & \textbf{0.253 } & \textbf{0.384 } & \textbf{0.264 } \\
    \bottomrule
    \end{tabular}%
    }
  \label{tbl:exp04_norm_layer_smaller}%
\end{table*}%

\begin{table*}[htbp]
  \centering
  \caption{The combined model with optimal architectures, which is compared to several existing LTSF models. All the results are averaged from four forecasting window length settings. Detailed results are presented in Table \ref{tbl:exp06_combination_model}. }
  \vspace{-0.7em}
  \resizebox{0.95\linewidth}{!}{
    \begin{tabular}{c|cc|cc|cc|cc|cc|cc|cc|cc|c}
    \toprule
    \multicolumn{1}{c|}{\textbf{dataset}} & \multicolumn{2}{c|}{\textbf{Illness}} & \multicolumn{2}{c|}{\textbf{ETTh1}} & \multicolumn{2}{c|}{\textbf{ETTh2}} & \multicolumn{2}{c|}{\textbf{ETTm1}} & \multicolumn{2}{c|}{\textbf{ETTm2}} & \multicolumn{2}{c|}{\textbf{Weather}} & \multicolumn{2}{c|}{\textbf{ECL}} & \multicolumn{2}{c|}{\textbf{Traffic}} & \multirow{2}[4]{*}{\textbf{1st Count}} \\
\cmidrule{1-17}    \textbf{model name} & \textbf{MSE} & \textbf{MAE} & \textbf{MSE} & \textbf{MAE} & \textbf{MSE} & \textbf{MAE} & \textbf{MSE} & \textbf{MAE} & \textbf{MSE} & \textbf{MAE} & \textbf{MSE} & \textbf{MAE} & \textbf{MSE} & \textbf{MAE} & \textbf{MSE} & \textbf{MAE} &  \\
    \midrule
    \midrule
    \textbf{Combined Model}
          & \textbf{1.278} & \textbf{0.740} & 0.410 & \textbf{0.433} & 0.335  & \textbf{0.377 } & 0.351  & \textbf{0.374} & 0.260  & 0.324  & \textbf{0.227} & \textbf{0.265} & \textbf{0.157} & \textbf{0.250} & \textbf{0.383} & \textbf{0.259} & 11 \\
    \midrule
    \textbf{FEDformer}
          & 2.597  & 1.070  & 0.428  & 0.454  & 0.388  & 0.434  & 0.382  & 0.422  & 0.410  & 0.420  & 0.310  & 0.357  & 0.207  & 0.321  & 0.604  & 0.372  & 0 \\
    \midrule 
    \textbf{PatchTST}
          & 1.480  & 0.807  & 0.413  & 0.434  & 0.331  & 0.381  & 0.353  & 0.382  & \textbf{0.257} & \textbf{0.317} & \textbf{0.227} & 0.266 & 0.159  & 0.253  & 0.391  & 0.264  & 3 \\
    \midrule
    \textbf{iTransformer}
          & 2.205  & 1.015  & 0.479  & 0.477  & 0.387  & 0.418  & 0.370  & 0.400  & 0.272  & 0.333  & 0.246  & 0.279  & 0.162  & 0.257  & 0.386  & 0.272  & 0 \\
    \midrule
    \textbf{CATS}
          & 3.784  & 1.342  & 0.415  & 0.434  & \textbf{0.330} & 0.381  & \textbf{0.344} & 0.379  & 0.258  & 0.323  & 0.228  & 0.267  & 0.159  & 0.251  & 0.385  & 0.261  & 2 \\
    \midrule
    \textbf{ARMA-Attention}
          & 1.989  & 0.942  & \textbf{0.406}  & \textbf{0.433} & 0.342  & 0.383  & 0.354  & 0.380  & 0.260  & 0.322  & 0.228  & 0.266  & 0.161  & 0.256  & 0.401  & 0.272  & 2 \\
    \bottomrule
    \end{tabular}%
    }
  \label{tbl:exp06_combination_model_smaller}%
\end{table*}%

\subsubsection{Forecasting Paradigm} \label{subsubsec:exp_forecasting_paradigm}
\
\par We then evaluate two forecasting paradigms in Figure \ref{fig:03_forecasting_objective}: direct-mapping and autoregressive. This experiment is conducted on decoder-only and encoder-decoder models, as models with uni-directional attention well suit the autoregressive paradigm.

\par The results in Table \ref{tbl:exp03_forecast_objective_smaller} show that \textbf{the direct-mapping paradigm significantly outperforms the autoregressive paradigm}, for both decoder-only and encoder-decoder models.

\begin{tcolorbox}[title = {Conclusion 3}] %
The direct-mapping forecasting paradigm significantly outperforms the autoregressive paradigm for LTSF.
\end{tcolorbox}

\par We attribute this to two main reasons. 
First, in the autoregressive paradigm, each predicted token is used as input for the next token, causing the ``\textbf{Error Accumulation}", where each step's error compounds in the subsequent steps, and results in explosive errors for long-term forecasting. 
Second, the teacher-forcing strategy accelerates training by feeding ground-truth values as inputs, enabling parallel loss computation.
However, this causes the ``\textbf{Exposure Bias}" problem, where the model must predict each token sequentially during inference, creating a mismatch between training and testing and degrading performance.
Furthermore, the rolling nature of the autoregressive paradigm makes it more time-consuming.

\subsubsection{Normalization Layer} \label{subsubsec:exp_normalization_layer}
\
\par Subsequently, we examine the impact of normalization layers, specifically LayerNorm (LN) and BatchNorm (BN).

\par The results in Table \ref{tbl:exp04_norm_layer_smaller} show that BN outperforms LN on some datasets (Illness, ETTh2, ETTm2, and Weather), whereas LN performs better on others (ETTh1, ETTm1, ECL, and Traffic). This suggests that the optimal choice of normalization layer may depend on dataset characteristics.

\par Upon analyzing these datasets, we find that the key distinction lies in the presence of outliers. Datasets like Illness, ETTh2, ETTm2, and Weather exhibit outliers and non-stationary patterns, while ETTh1, ETTm1, ECL, and Traffic are relatively stationary with fewer outliers.
To quantify the influence of outliers, we employ the Interquartile Range (IQR) method \cite{IQR_outlier}, identifying anomaly samples as those where the proportion of outliers outside the range $[Q1 - 1.5 \cdot IQR, Q3 + 1.5 \cdot IQR]$ exceeds a certain threshold (\eg, 5\%). Here $Q1$ and $Q3$ represent the first and third quartiles, and $IQR = Q3 - Q1$. 
We then calculate the ratio of anomaly samples to total samples for each dataset, as shown in the second line of Table \ref{tbl:exp04_norm_layer_smaller}.
The results indicate that BN performs better on datasets with a higher proportion of anomaly samples.

\par From these analyses, we conclude that:

\begin{tcolorbox}[title = {Conclusion 4}]
BatchNorm performs better for time series with more anomalies, while LayerNorm excels for more stationary time series containing fewer anomalies.
\end{tcolorbox}

\subsubsection{Forecasting length setting} \label{subsubsec:exp_forecasting_length_setting}
\
\par In this experiment, we switch to the variable forecasting length setting to validate our previous conclusions under different conditions, as described in Section \ref{subsec:experiment_setting}. Results are displayed in Table \ref{tbl:exp05_multi_pred_len_smaller} in the appendix due to page limits, with each model assigned a unique number for reference.

\par Firstly, comparing models No.2,3,4,5 confirms that bi-directional attention with joint-attention is more effective (Conclusion 1).
Next, comparing models No.1 and No.2 supports that more complete forecasting aggregation is beneficial (Conclusion 2).
Finally, comparing models No.4,5,6,7 verifies that the direct-mapping paradigm significantly outperforms the autoregressive paradigm (Conclusion 3). These analyses affirm the general applicability of our conclusions.

\begin{tcolorbox}[title = {Conclusion 5}]
Our conclusions hold for both fixed and variable forecasting length settings.
\end{tcolorbox}

\subsection{Combination of Optimal Architectures} 
\label{subsec:combination_model}
\par Based on the above conclusions, we construct an optimal Transformer architecture by combining the best choices, including bi-directional attention with joint-attention, complete forecasting aggregation, direct-mapping paradigm, and the BatchNorm layer.

\par We compare this combined model with several existing LTSF models, 
including FEDformer \cite{FEDformer}, PatchTST \cite{PatchTST}, iTransformer \cite{iTransformer}, CATS \cite{CATS}, and ARMA-Attention \cite{ARMA_Attention}, which cover diverse Transformer architectures listed in Table \ref{tbl:existing_models_taxonomy}, and are suitable baselines for comparison.
From the results reported in Table \ref{tbl:exp06_combination_model_smaller}, our combined model outperforms these models in most cases. A comparison with each model reveals the following:
\begin{itemize}[leftmargin=*]
    \item The encoder-decoder FEDformer model, despite incorporating time-series-specific designs like frequency-domain information capturing, underperforms due to its suboptimal attention mechanism, highlighting the importance of architectural design.
    \item The decoder-only ARMA-Attention model using uni-directional attention and the autoregressive paradigm, and the CATS model with uni-directional and cross-attention, are both less effective than our combined model, consistent with previous conclusions.
    \item The encoder-only PatchTST model, which applies partial forecasting aggregation, mainly focuses on tokens in the look-back window. In contrast, our model uses complete aggregation across both look-back and forecasting tokens, aligning with conclusion 2 and demonstrating that more complete aggregation improves performance. Similarly, the iTransformer model underperforms our model, due to its aggregation strategy.
\end{itemize}

\par These results not only validate our previous conclusions but also emphasize the power and significance of Transformer architectural designs for LTSF performance.

\section{Conclusions}

\par In this paper, we investigate the architectural variations in the Transformer-based models, to identify the optimal architectural designs for long-term time series forecasting. 
Specifically, we propose a taxonomy protocol that decouples these models across several key perspectives, including attention mechanisms, forecasting aggregation strategies, forecasting paradigms, and normalization layers.
Through extensive experiments on multiple long-term forecasting datasets, we derive several crucial conclusions: bi-directional attention with joint-attention is more effective, more complete and comprehensive forecasting aggregation leads to better performance, and the direct-mapping paradigm significantly outperforms the autoregressive approaches.
We hope this study provides valuable insights for future research on Transformer architectures for long-term time series forecasting.

\bibliographystyle{IEEEtran}
\bibliography{reference}

\newpage
\appendix

\numberwithin{equation}{section}
\section*{Appendix}

\section{Formulation of Transformers for LTSF} \label{sec:Transformer_formulation}
\par In the appendix, we briefly outline the formulation of primary components of Transformer blocks used in LTSF, as a supplement to Section \ref{subsec:transformer_arch_intro}.

\begin{itemize}[leftmargin=*]
    \item \textbf{Multi-head Self-Attention.} Given an input sequence $\mX \in \mathbb{R}^{L\times d_{model}}$, where $L$ is the sequence length and $d_{model}$ is the feature dimension, self-attention computes the query $\mathbf{Q}$, key $\mathbf{K}$, and value $\mathbf{V}$ matrices via learned projections:
    $$ \mathbf{Q} = \mathbf{X} \mathbf{W}_Q, \mathbf{K} = \mathbf{X} \mathbf{W}_K, \mathbf{V} = \mathbf{X} \mathbf{W}_V. $$
    where $\mathbf{W}_Q, \mathbf{W}_K \in \mathbb{R}^{d_{model} \times d_k}$ and $\mathbf{W}_V \in \mathbb{R}^{d_{model} \times d_v}$ are trainable parameters, with $d_k$ typically equals to $d_v$. The attention scores are calculated as:
    $$\mathrm{Attention}(\mathbf{Q},\mathbf{K},\mathbf{V})=\mathrm{softmax}\left(\frac{\mathbf{Q}\mathbf{K}^\top}{\sqrt{d_k}}\right)\mathbf{V}.$$
    Furthermore, multi-head self-attention (MHSA) extends this by applying $H$ parallel attention heads, concatenating their outputs, and projecting back to the original dimension:
    $$\mathrm{MHSA}(\mathbf{X})=\mathrm{Concat}(\mathrm{head}_1,\ldots,\mathrm{head}_H)\mathbf{W}_O,$$
    where each head $head_i$ is defined as:
    $$\mathrm{head}_i = \mathrm{Attention}( \mathbf{X} \mathbf{W}_Q^{(i)}, \mathbf{X} \mathbf{W}_K^{(i)}, \mathbf{X} \mathbf{W}_V^{(i)}). $$
    
    \item \textbf{Feed-Forward Neural Network.} After the self-attention, each time step in the sequence independently passes through a two-layer feed-forward neural network (FFN):
    $$\mathrm{FFN}(\mathbf{h}_t)=\sigma(\mathbf{h}_t\mathbf{W}_1+\mathbf{b}_1)\mathbf{W}_2+\mathbf{b}_2,$$
    where $\mathbf{W}_1, \mathbf{W}_2, \mathbf{b}_1, \mathbf{b}_2$ are learnable parameters, and $\sigma$ denotes a non-linear activation function such as ReLU.
    
    \item \textbf{Multi-head Cross-Attention.} 
    Cross-attention connects the encoder and decoder, aligning the target queries with source keys and values.
    For a query matrix $\mathbf{Q} \in \mathbb{R}^{T\times d_k}$ from decoder, a key matrix $\mathbf{K} \in \mathbb{R}^{L\times d_k}$ and a value matrix $\mathbf{V} \in \mathbb{R}^{T\times d_v}$ from encoder, multi-head cross-attention (MHCA) with $H$ heads is computed as:
    $$\mathrm{MHCA}(\mathbf{Q}, \mathbf{K}, \mathbf{V})=\mathrm{Concat}(\mathrm{head}_1,\ldots,\mathrm{head}_H)\mathbf{W}_O,$$
    $$\mathrm{head}_i = \mathrm{Attention}( \mathbf{Q} \mathbf{W}_Q^{(i)}, \mathbf{K} \mathbf{W}_K^{(i)}, \mathbf{V} \mathbf{W}_V^{(i)}).$$
    where $\mathbf{W}_Q^{(i)}, \mathbf{W}_K^{(i)}, \mathbf{W}_V^{(i)}$ are learnable weight matrices specific to the $i$-th head, and $\mathbf{W}_O$ is the projection matrix.
    
    \item \textbf{Prediction head.} The prediction head maps the latent embeddings from the encoder or decoder to the target values. Let $\mathbf{H} \in \mathbb{R}^{L\times d_{model}}$ be the latent embeddings and $\hat{\mX}_{t:t+T} \in \mathbb{R}^{T\times M}$ be the target values, the prediction head applies a transformation $f_{head}$ to produce predictions:
    $$\hat{\mX}_{t:t+T} = f_{head}(\mathbf{H}).$$
\end{itemize}

\section{Details of Experiments} 

\subsection{Datasets Details} \label{subsec:detailed_datasets}

\par We conduct our experiment on 8 popular datasets following previous researches \cite{Informer, Autoformer}. The statistics of these datasets are summarized in Table \ref{tbl:dataset_statistics}, and publicly available at \url{https://github.com/zhouhaoyi/Informer2020} and \url{https://github.com/thuml/Autoformer}.

\par \textbf{(1) ETTh1/ETTh2/ETTm1/ETTm2}. ETT dataset contains 7 indicators collected from electricity transformers from July 2016 to July 2018, including useful load, oil temperature, etc. Data points are recorded hourly for ETTh1 and ETTh2, while recorded every 15 minutes for ETTm1 and ETTm2.
\par \textbf{(2) Electricity}. Electricity dataset contains the hourly electricity consumption (in KWh) of 321 customers from 2012 to 2014.
\par \textbf{(3) Traffic}. Traffic dataset contains hourly road occupancy rate data measured by different sensors on San Francisco Bay area freeways in 2 years. The data is from the California Department of Transportation.
\par \textbf{(4) Illness}. Illness dataset includes 7 weekly recorded indicators of influenza-like illness patients data from Centers for
Disease Control and Prevention of the United States between 2002 and 2021.
\par \textbf{(5) Weather}. Weather dataset contains 21 meteorological indicators, like temperature, humidity, etc. The dataset is recorded every 10 minutes for the 2020 whole year.

\begin{table}[t]
  \centering
  \caption{The statistics of 8 datasets, including number of variates, total time steps, and frequency of data sampling.}
  \vspace{-0.7em}
  \resizebox{\linewidth}{!}{
    \begin{tabular}{ccccccccc}
    \toprule
    \textbf{Dataset} & \textbf{ETTh1} & \textbf{ETTh2} & \textbf{ETTm1} & \textbf{ETTm2} & \textbf{Electricity} & \textbf{Traffic} & \textbf{Weather} & \textbf{Illness} \\
    \midrule
    \midrule
    \textbf{Variates} & 7     & 7     & 7     & 7     & 321   & 862   & 21    & 7 \\
    \textbf{Time steps} & 17,420 & 17,420 & 69,680 & 69,680 & 26,304 & 17,544 & 52,695 & 966 \\
    \textbf{Frequency} & 1hour & 1hour & 15min & 15min & 1hour & 1hour & 10min & 1week \\
    \bottomrule
    \end{tabular}%
  }
  \label{tbl:dataset_statistics}%
\end{table}%

\subsection{Hyperparameters Details} \label{subsec:detailed_hyperparams}

\par Detailed hyperparameters for our experiments are provided in Table \ref{tbl:main_parameters}. Consistent hyperparameter settings guarantee fair comparisons between different architectures.

\begin{table}[t]
\caption{Main hyperparameters for different datasets.}
\label{tbl:main_parameters}
\vspace{-0.7em}
\begin{center}
\begin{small}
\resizebox{0.95\linewidth}{!}{
\begin{tabular}{lcc} 
 \toprule
 & \multicolumn{2}{c}{\sc{Hyperparameters for different datasets}} \\
 & Illness & Others (w/o Illness) \\
 \midrule
 Look-back window & 120 & 512 \\
 Forecasting window & \{24, 36, 48, 60\} & \{96, 192, 336, 720\} \\
 Patch length & 6 & 16 \\
 Patch stride & 6 & 16 \\
 Stacked Layers & \multicolumn{2}{c}{6 (joint-attention) or 3+3 (cross-attention)} \\
 Embedding dimension & \multicolumn{2}{c}{512} \\
 Attention heads & \multicolumn{2}{c}{8} \\
 Positional embedding & \multicolumn{2}{c}{learnable} \\
 Learning rate & \multicolumn{2}{c}{0.001} \\
 Optimizer & \multicolumn{2}{c}{Adam} \\
 
 \bottomrule
\end{tabular}
}
\end{small}
\end{center}
\end{table}

\subsection{Variable Forecasting Length Results}
\par The results for the experiments of variable forecasting length setting in Section \ref{subsubsec:exp_forecasting_length_setting} are reported in Table \ref{tbl:exp05_multi_pred_len_smaller} in this section, due to the page limit in the main text.

\begin{table*}[htbp]
  \centering
  \caption{Experiments on variable forecasting length setting. For simplicity, we use these abbreviations in model names: ``aggr." for ``aggregation”, ``direct." for ``direct-mapping", and ``autoreg." for the ``autoregressive" paradigm. All the results are averaged from four forecasting window length settings. Detailed results are presented in Table \ref{tbl:exp05_multi_pred_len} in the appendix. }
  \vspace{-0.7em}
    \resizebox{0.92\linewidth}{!}{
    \begin{tabular}{c|c|cc|cc|cc|cc|cc|cc|cc|cc}
    \toprule
    \multicolumn{2}{c|}{\textbf{dataset}} & \multicolumn{2}{c|}{\textbf{Illness}} & \multicolumn{2}{c|}{\textbf{ETTh1}} & \multicolumn{2}{c|}{\textbf{ETTh2}} & \multicolumn{2}{c|}{\textbf{ETTm1}} & \multicolumn{2}{c|}{\textbf{ETTm2}} & \multicolumn{2}{c|}{\textbf{Weather}} & \multicolumn{2}{c|}{\textbf{ECL}} & \multicolumn{2}{c}{\textbf{Traffic}} \\
    \midrule
    \textbf{No.} & \textbf{model} & \textbf{MSE} & \textbf{MAE} & \textbf{MSE} & \textbf{MAE} & \textbf{MSE} & \textbf{MAE} & \textbf{MSE} & \textbf{MAE} & \textbf{MSE} & \textbf{MAE} & \textbf{MSE} & \textbf{MAE} & \textbf{MSE} & \textbf{MAE} & \textbf{MSE} & \textbf{MAE} \\
    \midrule
    \midrule
    \textbf{1} & \makecell{\textbf{Encoder-only +} \\ \textbf{complete aggr. + direct.}}
          & \textbf{1.345 } & \textbf{0.770 } & \textbf{0.419 } & \textbf{0.440 } & \textbf{0.357 } & \textbf{0.395 } & \textbf{0.363 } & \textbf{0.391 } & \textbf{0.265 } & \textbf{0.325 } & \textbf{0.231 } & \textbf{0.270 } & 0.161  & 0.258  & 0.393  & 0.276  \\
    \midrule
    \textbf{2} & \makecell{\textbf{Encoder-only +} \\ \textbf{no aggr. + direct.}}
          & 1.429  & 0.791  & 0.479  & 0.453  & 0.431  & 0.449  & 0.377  & 0.399  & 0.306  & 0.364  & 0.247  & 0.284  & \textbf{0.160 } & \textbf{0.253 } & \textbf{0.386 } & 0.265  \\
    \midrule
    \textbf{3} & \makecell{\textbf{Prefix decoder +} \\ \textbf{no aggr. + direct.}}
          & 1.466  & 0.801  & 0.505  & 0.469  & 0.415  & 0.448  & 0.455  & 0.445  & 0.320  & 0.380  & 0.247  & 0.296  & 0.164  & 0.260  & \textbf{0.386 } & 0.263  \\
    \midrule
    \textbf{4} & \makecell{\textbf{Encoder-decoder +} \\ \textbf{no aggr. + direct.}}
          & 1.838  & 0.817  & 0.518  & 0.469  & 0.394  & 0.431  & 0.418  & 0.418  & 0.302  & 0.361  & 0.238  & 0.280  & 0.162  & 0.254  & 0.389  & 0.265  \\
    \midrule
   \textbf{5} & \makecell{\textbf{Decoder-only +} \\ \textbf{no aggr. + direct.}}
          & 2.122  & 0.954  & 0.557  & 0.509  & 0.481  & 0.476  & 0.436  & 0.423  & 0.331  & 0.382  & 0.256  & 0.302  & 0.163  & 0.259  & \textbf{0.386 } & \textbf{0.262 } \\
    \midrule
    \textbf{6} & \makecell{\textbf{Encoder-decoder +} \\ \textbf{no aggr. + autoerg.}}
          & 4.394  & 1.535  & 1.319  & 0.722  & 0.487  & 0.494  & 1.128  & 0.665  & 0.549  & 0.480  & 0.525  & 0.444  & 0.439  & 0.400  & 4.394  & 1.535  \\
    \midrule
    \textbf{7} & \makecell{\textbf{Decoder-only +} \\ \textbf{no aggr. + autoreg.}}
          & 5.318  & 1.652  & 1.233  & 0.766  & 0.622  & 0.540  & 1.352  & 0.749  & 0.598  & 0.511  & 0.543  & 0.461  & 0.465  & 0.410  & 5.318  & 1.652  \\
    \bottomrule
    \end{tabular}%
    }
  \label{tbl:exp05_multi_pred_len_smaller}%
\end{table*}%

\subsection{Detailed Experiments} \label{subsec:detailed_exps}

\subsubsection{Full Table of Attention Mechanism}
\par This section includes the full table of experiments on the attention mechanisms, as an extended version of Table \ref{tbl:exp01_attention_smaller} in Section \ref{subsubsec:exp_attention_mechanism}. 
Results are reported in Table \ref{tbl:exp01_attention}.

\begin{table*}[htbp]
  \centering
  \caption{Full table on six attention mechanisms. In our notation, ``24/96" indicates a forecasting window length of 24 for the Illness dataset and 96 for others, with the same logic applying to ``36/192", ``48/336" and ``60/720". ``avg" denotes the average of four forecasting window lengths.}
  \vspace{-0.7em}
  \resizebox{0.95\linewidth}{!}{
    \begin{tabular}{c|c|cc|cc|cc|cc|cc|cc|cc|cc|c}
    \toprule
    \multicolumn{2}{c|}{\textbf{dataset}} & \multicolumn{2}{c|}{\textbf{Illness}} & \multicolumn{2}{c|}{\textbf{ETTh1}} & \multicolumn{2}{c|}{\textbf{ETTh2}} & \multicolumn{2}{c|}{\textbf{ETTm1}} & \multicolumn{2}{c|}{\textbf{ETTm2}} & \multicolumn{2}{c|}{\textbf{Weather}} & \multicolumn{2}{c|}{\textbf{ECL}} & \multicolumn{2}{c|}{\textbf{Traffic}} & \multirow{2}[4]{*}{\textbf{1st Count}} \\
\cmidrule{1-18}    \textbf{model} & \textbf{pred\_len} & \textbf{MSE} & \textbf{MAE} & \textbf{MSE} & \textbf{MAE} & \textbf{MSE} & \textbf{MAE} & \textbf{MSE} & \textbf{MAE} & \textbf{MSE} & \textbf{MAE} & \textbf{MSE} & \textbf{MAE} & \textbf{MSE} & \textbf{MAE} & \textbf{MSE} & \textbf{MAE} &  \\
    \midrule
    \midrule
    \multirow{5}[1]{*}{\textbf{Encoder-only}} & \textbf{24/96} & \textbf{1.274 } & \textbf{0.726 } & \textbf{0.415 } & \textbf{0.416 } & 0.303  & 0.361  & 0.319  & 0.344  & 0.188  & \textbf{0.266 } & 0.146  & 0.199  & 0.128  & 0.220  & 0.350  & 0.240  & \multirow{5}[1]{*}{38} \\
          & \textbf{36/192} & 1.240  & 0.735  & \textbf{0.506 } & \textbf{0.448 } & 0.367  & \textbf{0.399 } & 0.351  & 0.373  & 0.278  & 0.340  & \textbf{0.190 } & \textbf{0.238 } & 0.146  & 0.241  & 0.375  & 0.249  &  \\
          & \textbf{48/336} & \textbf{1.320 } & \textbf{0.771 } & \textbf{0.449 } & \textbf{0.427 } & 0.407  & 0.450  & \textbf{0.367 } & \textbf{0.394 } & \textbf{0.304 } & \textbf{0.360 } & \textbf{0.250 } & \textbf{0.284 } & \textbf{0.162 } & \textbf{0.256 } & \textbf{0.395 } & \textbf{0.271 } &  \\
          & \textbf{60/720} & \textbf{1.534 } & 0.830  & 0.525  & 0.502  & 0.493  & 0.496  & \textbf{0.415 } & \textbf{0.421 } & 0.390  & 0.412  & 0.345  & 0.349  & \textbf{0.197 } & \textbf{0.287 } & 0.420  & 0.290  &  \\
          & \textbf{avg} & \textbf{1.342 } & \textbf{0.766 } & \textbf{0.474 } & \textbf{0.448 } & 0.393  & 0.427  & \textbf{0.363 } & \textbf{0.383 } & 0.290  & 0.345  & 0.233  & \textbf{0.268 } & \textbf{0.158 } & \textbf{0.251 } & 0.385  & 0.263  &  \\
    \midrule
    \multirow{5}[0]{*}{\textbf{Prefix decoder}} & \textbf{24/96} & 1.301  & 0.755  & 0.439  & 0.421  & 0.318  & 0.386  & 0.313  & 0.363  & 0.203  & 0.299  & 0.149  & 0.204  & \textbf{0.127 } & \textbf{0.219 } & \textbf{0.348 } & \textbf{0.238 } & \multirow{5}[0]{*}{20} \\
          & \textbf{36/192} & \textbf{1.154 } & \textbf{0.710 } & 0.509  & 0.464  & 0.383  & 0.413  & 0.381  & 0.396  & 0.263  & 0.338  & 0.205  & 0.253  & \textbf{0.145 } & \textbf{0.238 } & \textbf{0.370 } & \textbf{0.246 } &  \\
          & \textbf{48/336} & 1.582  & 0.836  & 0.489  & 0.448  & 0.402  & 0.451  & 0.409  & 0.421  & 0.344  & 0.393  & 0.277  & 0.309  & \textbf{0.162 } & \textbf{0.256 } & 0.396  & 0.272  &  \\
          & \textbf{60/720} & 1.558  & \textbf{0.821 } & 0.563  & 0.530  & 0.456  & 0.481  & 0.519  & 0.458  & 0.413  & 0.431  & 0.328  & 0.349  & \textbf{0.197 } & 0.290  & \textbf{0.418 } & \textbf{0.288 } &  \\
          & \textbf{avg} & 1.399  & 0.781  & 0.520  & 0.468  & 0.390  & 0.433  & 0.406  & 0.410  & 0.306  & 0.365  & 0.240  & 0.279  & \textbf{0.158 } & \textbf{0.251 } & \textbf{0.383 } & \textbf{0.261 } &  \\
    \midrule
    \multirow{5}[0]{*}{\textbf{Decoder-only}} & \textbf{24/96} & 2.124  & 0.956  & 0.452  & 0.441  & 0.390  & 0.425  & \textbf{0.310 } & \textbf{0.341 } & 0.244  & 0.307  & 0.152  & 0.205  & 0.128  & 0.221  & 0.350  & 0.244  & \multirow{5}[0]{*}{4} \\
          & \textbf{36/192} & 2.241  & 0.987  & 0.547  & 0.476  & 0.417  & 0.445  & \textbf{0.341 } & \textbf{0.371 } & 0.273  & 0.343  & 0.207  & 0.260  & 0.147  & 0.242  & 0.371  & 0.249  &  \\
          & \textbf{48/336} & 1.943  & 0.932  & 0.594  & 0.515  & 0.437  & 0.481  & 0.391  & 0.407  & 0.366  & 0.403  & 0.273  & 0.303  & 0.163  & 0.258  & 0.396  & 0.272  &  \\
          & \textbf{60/720} & 2.187  & 0.981  & 0.615  & 0.586  & 0.520  & 0.504  & 0.560  & 0.476  & 0.413  & 0.427  & 0.349  & 0.356  & 0.199  & 0.290  & 0.420  & 0.290  &  \\
          & \textbf{avg} & 2.124  & 0.964  & 0.552  & 0.505  & 0.441  & 0.464  & 0.401  & 0.399  & 0.324  & 0.370  & 0.245  & 0.281  & 0.159  & 0.253  & 0.384  & 0.264  &  \\
    \midrule
    \multirow{5}[0]{*}{\textbf{Double-encoder}} & \textbf{24/96} & 2.386  & 1.058  & 0.582  & 0.499  & 0.342  & 0.397  & 0.424  & 0.390  & 0.289  & 0.353  & 0.267  & 0.316  & 0.173  & 0.271  & 0.383  & 0.297  & \multirow{5}[0]{*}{6} \\
          & \textbf{36/192} & 2.333  & 0.992  & 0.597  & 0.516  & \textbf{0.361 } & 0.410  & 0.433  & 0.395  & 0.313  & 0.365  & 0.291  & 0.332  & 0.181  & 0.273  & 0.395  & 0.299  &  \\
          & \textbf{48/336} & 2.214  & 0.985  & 0.585  & 0.526  & \textbf{0.327 } & \textbf{0.411 } & 0.444  & 0.401  & 0.353  & 0.387  & 0.319  & 0.347  & 0.192  & 0.277  & 0.408  & 0.301  &  \\
          & \textbf{60/720} & 2.250  & 1.015  & 0.601  & 0.572  & \textbf{0.379 } & \textbf{0.427 } & 0.463  & 0.435  & 0.418  & 0.424  & 0.362  & 0.375  & 0.200  & 0.290  & 0.430  & 0.304  &  \\
          & \textbf{avg} & 2.296  & 1.013  & 0.591  & 0.528  & \textbf{0.359 } & 0.418  & 0.391  & 0.400  & 0.343  & 0.382  & 0.310  & 0.343  & 0.187  & 0.278  & 0.404  & 0.300  &  \\
    \midrule
    \multirow{5}[0]{*}{\textbf{Encoder-decoder}} & \textbf{24/96} & 1.899  & 0.807  & 0.506  & 0.454  & \textbf{0.299 } & \textbf{0.347 } & 0.313  & 0.342  & 0.190  & 0.268  & \textbf{0.144 } & \textbf{0.193 } & 0.128  & 0.221  & 0.351  & 0.242  & \multirow{5}[0]{*}{11} \\
          & \textbf{36/192} & 1.941  & 0.851  & 0.509  & 0.455  & 0.384  & 0.405  & 0.377  & 0.396  & \textbf{0.261 } & \textbf{0.314 } & 0.196  & 0.250  & 0.147  & 0.239  & 0.378  & 0.248  &  \\
          & \textbf{48/336} & 1.564  & 0.781  & 0.490  & 0.467  & 0.344  & 0.418  & 0.381  & 0.409  & 0.310  & 0.364  & 0.259  & 0.298  & 0.165  & 0.257  & 0.398  & 0.274  &  \\
          & \textbf{60/720} & 1.935  & 0.843  & 0.533  & 0.476  & 0.410  & 0.448  & 0.536  & 0.473  & 0.387  & \textbf{0.407 } & 0.318  & 0.340  & 0.198  & 0.291  & 0.422  & 0.292  &  \\
          & \textbf{avg} & 1.835  & 0.816  & 0.510  & 0.463  & \textbf{0.359 } & 0.406  & 0.402  & 0.405  & \textbf{0.287 } & \textbf{0.338 } & \textbf{0.229 } & 0.270  & 0.160  & 0.252  & 0.387  & 0.264  &  \\
    \midrule
    \multirow{5}[1]{*}{\textbf{Double-decoder }} & \textbf{24/96} & 1.377  & 0.772  & 0.461  & 0.426  & 0.308  & 0.352  & 0.329  & 0.361  & \textbf{0.187 } & 0.271  & 0.152  & 0.201  & 0.128  & 0.221  & 0.358  & 0.244  & \multirow{5}[1]{*}{8} \\
          & \textbf{36/192} & 1.517  & 0.830  & 0.583  & 0.468  & 0.399  & 0.409  & 0.352  & 0.377  & 0.274  & 0.320  & 0.195  & 0.243  & 0.146  & 0.239  & 0.378  & 0.248  &  \\
          & \textbf{48/336} & 1.460  & 0.820  & 0.462  & 0.435  & 0.359  & 0.415  & 0.501  & 0.465  & 0.309  & 0.363  & 0.252  & 0.299  & 0.163  & 0.259  & 0.398  & 0.274  &  \\
          & \textbf{60/720} & 1.541  & 0.863  & \textbf{0.504 } & \textbf{0.473 } & 0.407  & 0.448  & 0.485  & 0.454  & \textbf{0.386 } & 0.410  & \textbf{0.317 } & \textbf{0.337 } & 0.198  & 0.288  & 0.422  & 0.292  &  \\
          & \textbf{avg} & 1.474  & 0.821  & 0.503  & 0.451  & 0.361  & \textbf{0.399 } & 0.417  & 0.414  & 0.289  & 0.341  & \textbf{0.229 } & 0.270  & 0.159  & 0.252  & 0.389  & 0.265  &  \\
    \bottomrule
    \end{tabular}%
    }
  \label{tbl:exp01_attention}%
\end{table*}%

\subsubsection{Full Table of Forecasting Aggregation Approaches}
\par This section includes the full table of experiments on the forecasting aggregation approaches, as an extended version of Table \ref{tbl:exp02_feature_fusion_smaller} in Section \ref{subsubsec:exp_forecasting_aggregation}. 
Results are reported in Table \ref{tbl:exp02_feature_fusion}.

\begin{table*}[htbp]
  \centering
  \caption{Full table on three forecasting aggregation methods.}
  \vspace{-0.7em}
    \resizebox{0.95\linewidth}{!}{
    \begin{tabular}{c|c|cc|cc|cc|cc|cc|cc|cc|cc|c}
    \toprule
    \multicolumn{2}{c|}{\textbf{dataset}} & \multicolumn{2}{c|}{\textbf{Illness}} & \multicolumn{2}{c|}{\textbf{ETTh1}} & \multicolumn{2}{c|}{\textbf{ETTh2}} & \multicolumn{2}{c|}{\textbf{ETTm1}} & \multicolumn{2}{c|}{\textbf{ETTm2}} & \multicolumn{2}{c|}{\textbf{Weather}} & \multicolumn{2}{c|}{\textbf{ECL}} & \multicolumn{2}{c|}{\textbf{Traffic}} & \multirow{2}[4]{*}{\textbf{1st Count}} \\
\cmidrule{1-18}    \textbf{model} & \textbf{pred\_len} & \textbf{MSE} & \textbf{MAE} & \textbf{MSE} & \textbf{MAE} & \textbf{MSE} & \textbf{MAE} & \textbf{MSE} & \textbf{MAE} & \textbf{MSE} & \textbf{MAE} & \textbf{MSE} & \textbf{MAE} & \textbf{MSE} & \textbf{MAE} & \textbf{MSE} & \textbf{MAE} &  \\
    \midrule
    \midrule
    \multirow{5}[2]{*}{\textbf{No aggregation}} & \textbf{24/96} & 1.274  & 0.726  & 0.415  & 0.416  & 0.303  & 0.361  & 0.319  & 0.344  & 0.188  & 0.266  & 0.146  & 0.199  & \textbf{0.128 } & \textbf{0.220 } & \textbf{0.350 } & \textbf{0.240 } & \multirow{5}[2]{*}{6} \\
          & \textbf{36/192} & 1.240  & 0.735  & 0.506  & 0.448  & 0.367  & 0.399  & 0.351  & 0.373  & 0.278  & 0.340  & \textbf{0.190 } & \textbf{0.238 } & 0.146  & 0.241  & 0.375  & 0.249  &  \\
          & \textbf{48/336} & 1.420  & 0.811  & 0.449  & 0.435  & 0.407  & 0.450  & 0.367  & 0.394  & 0.304  & 0.360  & 0.250  & 0.284  & 0.163  & 0.257  & 0.395  & 0.271  &  \\
          & \textbf{60/720} & 1.534  & 0.830  & 0.525  & 0.502  & 0.493  & 0.496  & 0.415  & 0.421  & 0.390  & 0.412  & 0.345  & 0.349  & 0.197  & 0.287  & 0.420  & 0.290  &  \\
          & \textbf{avg} & 1.367  & 0.776  & 0.474  & 0.450  & 0.393  & 0.427  & 0.363  & 0.383  & 0.290  & 0.345  & 0.233  & 0.268  & 0.159  & 0.251  & 0.385  & 0.263  &  \\
    \midrule
    \multirow{5}[2]{*}{\textbf{Partial aggregation}} & \textbf{24/96} & 1.290  & 0.733  & 0.397  & 0.422  & 0.285  & 0.343  & 0.301  & 0.365  & 0.169  & 0.268  & 0.153  & 0.206  & \textbf{0.128 } & \textbf{0.220 } & 0.352  & 0.240  & \multirow{5}[2]{*}{10} \\
          & \textbf{36/192} & 1.191  & 0.715  & 0.444  & 0.445  & 0.346  & 0.385  & 0.347  & 0.394  & 0.235  & 0.312  & 0.205  & 0.255  & \textbf{0.144 } & \textbf{0.239 } & 0.378  & 0.243  &  \\
          & \textbf{48/336} & 1.413  & 0.803  & 0.450  & 0.427  & 0.361  & 0.399  & \textbf{0.362 } & \textbf{0.374 } & 0.292  & 0.351  & 0.269  & 0.300  & \textbf{0.162 } & \textbf{0.256 } & 0.398  & 0.274  &  \\
          & \textbf{60/720} & 1.461  & 0.813  & 0.463  & 0.479  & 0.408  & 0.442  & \textbf{0.408 } & \textbf{0.412 } & 0.383  & 0.406  & 0.325  & 0.348  & 0.199  & 0.289  & 0.420  & 0.290  &  \\
          & \textbf{avg} & 1.339  & 0.766  & 0.439  & 0.443  & 0.350  & 0.392  & 0.355  & 0.386  & 0.270  & 0.334  & 0.238  & 0.277  & 0.158  & 0.251  & 0.387  & 0.262  &  \\
    \midrule
    \multirow{5}[2]{*}{\textbf{Complete aggregation}} & \textbf{24/96} & \textbf{1.156 } & \textbf{0.697 } & \textbf{0.367 } & \textbf{0.402 } & \textbf{0.275 } & \textbf{0.336 } & \textbf{0.294 } & \textbf{0.338 } & \textbf{0.163 } & \textbf{0.261 } & \textbf{0.145 } & \textbf{0.197 } & \textbf{0.128 } & \textbf{0.220 } & \textbf{0.350 } & \textbf{0.240 } & \multirow{5}[2]{*}{72} \\
          & \textbf{36/192} & \textbf{1.096 } & \textbf{0.691 } & \textbf{0.413 } & \textbf{0.428 } & \textbf{0.337 } & \textbf{0.375 } & \textbf{0.336 } & \textbf{0.368 } & \textbf{0.223 } & \textbf{0.301 } & 0.193  & 0.243  & 0.145  & 0.240  & \textbf{0.374 } & \textbf{0.247 } &  \\
          & \textbf{48/336} & \textbf{1.407 } & \textbf{0.802 } & \textbf{0.419 } & \textbf{0.437 } & \textbf{0.343 } & \textbf{0.379 } & 0.363  & 0.375  & \textbf{0.279 } & \textbf{0.335 } & \textbf{0.245 } & \textbf{0.282 } & \textbf{0.162 } & \textbf{0.256 } & \textbf{0.390 } & \textbf{0.265 } &  \\
          & \textbf{60/720} & \textbf{1.453 } & \textbf{0.771 } & \textbf{0.442 } & \textbf{0.465 } & \textbf{0.383 } & \textbf{0.418 } & 0.410  & 0.415  & \textbf{0.375 } & \textbf{0.397 } & \textbf{0.324 } & \textbf{0.338 } & \textbf{0.194 } & \textbf{0.285 } & \textbf{0.417 } & \textbf{0.284 } &  \\
          & \textbf{avg} & \textbf{1.278 } & \textbf{0.740 } & \textbf{0.410 } & \textbf{0.433 } & \textbf{0.335 } & \textbf{0.377 } & \textbf{0.351 } & \textbf{0.374 } & \textbf{0.260 } & \textbf{0.324 } & \textbf{0.227 } & \textbf{0.265 } & \textbf{0.157 } & \textbf{0.250 } & \textbf{0.383 } & \textbf{0.259 } &  \\
    \bottomrule
    \end{tabular}%
    }
  \label{tbl:exp02_feature_fusion}%
\end{table*}%

\subsubsection{Full Table of Forecasting Paradigm}
\par This section includes the full table of experiments on the forecasting paradigms, as an extended version of Table \ref{tbl:exp03_forecast_objective_smaller} in Section \ref{subsubsec:exp_forecasting_paradigm}. 
Results are reported in Table \ref{tbl:exp03_forecast_objective}.

\begin{table*}[htbp]
  \centering
  \caption{Full table on forecasting paradigms for decoder-only and encoder-decoder models.}
  \vspace{-0.7em}
    \resizebox{0.92\linewidth}{!}{
    \begin{tabular}{c|c|cc|cc|cc|cc|cc|cc|cc|cc}
    \toprule
    \multicolumn{2}{c|}{\textbf{dataset}} & \multicolumn{2}{c|}{\textbf{Illness}} & \multicolumn{2}{c|}{\textbf{ETTh1}} & \multicolumn{2}{c|}{\textbf{ETTh2}} & \multicolumn{2}{c|}{\textbf{ETTm1}} & \multicolumn{2}{c|}{\textbf{ETTm2}} & \multicolumn{2}{c|}{\textbf{Weather}} & \multicolumn{2}{c|}{\textbf{ECL}} & \multicolumn{2}{c}{\textbf{Traffic}} \\
    \midrule
    \textbf{model} & \textbf{pred\_len} & \textbf{MSE} & \textbf{MAE} & \textbf{MSE} & \textbf{MAE} & \textbf{MSE} & \textbf{MAE} & \textbf{MSE} & \textbf{MAE} & \textbf{MSE} & \textbf{MAE} & \textbf{MSE} & \textbf{MAE} & \textbf{MSE} & \textbf{MAE} & \textbf{MSE} & \textbf{MAE} \\
    \midrule
    \midrule
    \multirow{5}[1]{*}{\makecell{\textbf{Decoder-} \\ \textbf{direct}}} & \textbf{24/96} & \textbf{2.124 } & \textbf{0.956 } & \textbf{0.594 } & \textbf{0.515 } & \textbf{0.390 } & \textbf{0.425 } & \textbf{0.310 } & \textbf{0.341 } & \textbf{0.244 } & \textbf{0.307 } & \textbf{0.152 } & \textbf{0.205 } & \textbf{0.128 } & \textbf{0.221 } & \textbf{0.350 } & \textbf{0.244 } \\
          & \textbf{36/192} & \textbf{2.241 } & \textbf{0.987 } & \textbf{0.547 } & \textbf{0.476 } & \textbf{0.417 } & \textbf{0.445 } & \textbf{0.341 } & \textbf{0.371 } & \textbf{0.273 } & \textbf{0.343 } & \textbf{0.207 } & \textbf{0.260 } & \textbf{0.147 } & \textbf{0.242 } & \textbf{0.371 } & \textbf{0.249 } \\
          & \textbf{48/336} & \textbf{1.943 } & \textbf{0.932 } & \textbf{0.452 } & \textbf{0.441 } & \textbf{0.437 } & \textbf{0.481 } & \textbf{0.391 } & \textbf{0.407 } & \textbf{0.366 } & \textbf{0.403 } & \textbf{0.273 } & \textbf{0.303 } & \textbf{0.163 } & \textbf{0.258 } & \textbf{0.396 } & \textbf{0.272 } \\
          & \textbf{60/720} & \textbf{2.187 } & \textbf{0.981 } & \textbf{0.615 } & \textbf{0.586 } & \textbf{0.520 } & \textbf{0.504 } & \textbf{0.560 } & \textbf{0.476 } & \textbf{0.413 } & \textbf{0.427 } & \textbf{0.349 } & \textbf{0.356 } & \textbf{0.199 } & \textbf{0.290 } & \textbf{0.420 } & \textbf{0.290 } \\
          & \textbf{avg} & \textbf{2.124 } & \textbf{0.964 } & \textbf{0.552 } & \textbf{0.505 } & \textbf{0.441 } & \textbf{0.464 } & \textbf{0.401 } & \textbf{0.399 } & \textbf{0.324 } & \textbf{0.370 } & \textbf{0.245 } & \textbf{0.281 } & \textbf{0.159 } & \textbf{0.253 } & \textbf{0.384 } & \textbf{0.264 } \\
    \midrule
    \multirow{5}[1]{*}{\makecell{\textbf{Decoder-} \\ \textbf{autoregressive}}} & \textbf{24/96} & 5.795  & 1.780  & 1.012  & 0.749  & 0.604  & 0.527  & 1.321  & 0.735  & 0.542  & 0.483  & 0.502  & 0.443  & 0.432  & 0.398  & 5.795  & 1.780  \\
          & \textbf{36/192} & 5.156  & 1.615  & 1.234  & 0.760  & 0.612  & 0.531  & 1.344  & 0.742  & 0.601  & 0.512  & 0.531  & 0.451  & 0.461  & 0.408  & 5.156  & 1.615  \\
          & \textbf{48/336} & 5.070  & 1.588  & 1.338  & 0.768  & 0.663  & 0.558  & 1.378  & 0.760  & 0.621  & 0.521  & 0.555  & 0.462  & 0.471  & 0.412  & 5.070  & 1.588  \\
          & \textbf{60/720} & 5.251  & 1.624  & 1.349  & 0.787  & 0.608  & 0.545  & 1.363  & 0.759  & 0.629  & 0.527  & 0.584  & 0.489  & 0.494  & 0.421  & 5.251  & 1.624  \\
          & \textbf{avg} & 5.318  & 1.652  & 1.233  & 0.766  & 0.622  & 0.540  & 1.352  & 0.749  & 0.598  & 0.511  & 0.543  & 0.461  & 0.465  & 0.410  & 5.318  & 1.652  \\
    \midrule
    \midrule
    \multirow{5}[1]{*}{\makecell{\textbf{Encoder-decoder-} \\ \textbf{direct}}} & \textbf{24/96} & \textbf{1.899 } & \textbf{0.807 } & \textbf{0.506 } & \textbf{0.454 } & \textbf{0.299 } & \textbf{0.347 } & \textbf{0.313 } & \textbf{0.342 } & \textbf{0.190 } & \textbf{0.268 } & \textbf{0.144 } & \textbf{0.193 } & \textbf{0.128 } & \textbf{0.221 } & \textbf{0.351 } & \textbf{0.242 } \\
          & \textbf{36/192} & \textbf{1.941 } & \textbf{0.851 } & \textbf{0.509 } & \textbf{0.455 } & \textbf{0.384 } & \textbf{0.405 } & \textbf{0.377 } & \textbf{0.396 } & \textbf{0.261 } & \textbf{0.314 } & \textbf{0.196 } & \textbf{0.250 } & \textbf{0.147 } & \textbf{0.239 } & \textbf{0.378 } & \textbf{0.248 } \\
          & \textbf{48/336} & \textbf{1.564 } & \textbf{0.781 } & \textbf{0.490 } & \textbf{0.467 } & \textbf{0.344 } & \textbf{0.418 } & \textbf{0.381 } & \textbf{0.409 } & \textbf{0.310 } & \textbf{0.364 } & \textbf{0.259 } & \textbf{0.298 } & \textbf{0.165 } & \textbf{0.257 } & \textbf{0.398 } & \textbf{0.274 } \\
          & \textbf{60/720} & \textbf{1.935 } & \textbf{0.843 } & \textbf{0.533 } & \textbf{0.476 } & \textbf{0.410 } & \textbf{0.448 } & \textbf{0.536 } & \textbf{0.473 } & \textbf{0.387 } & \textbf{0.407 } & \textbf{0.318 } & \textbf{0.340 } & \textbf{0.198 } & \textbf{0.291 } & \textbf{0.422 } & \textbf{0.292 } \\
          & \textbf{avg} & \textbf{1.835 } & \textbf{0.821 } & \textbf{0.510 } & \textbf{0.463 } & \textbf{0.359 } & \textbf{0.405 } & \textbf{0.402 } & \textbf{0.405 } & \textbf{0.287 } & \textbf{0.338 } & \textbf{0.229 } & \textbf{0.270 } & \textbf{0.160 } & \textbf{0.252 } & \textbf{0.387 } & \textbf{0.264 } \\
    \midrule
    \multirow{5}[1]{*}{\makecell{\textbf{Encoder-decoder-} \\ \textbf{autoregressive}}} & \textbf{24/96} & 4.589  & 1.598  & 1.150  & 0.698  & 0.418  & 0.451  & 1.115  & 0.645  & 0.509  & 0.455  & 0.481  & 0.420  & 0.408  & 0.386  & 4.589  & 1.598  \\
          & \textbf{36/192} & 4.405  & 1.538  & 1.301  & 0.721  & 0.482  & 0.487  & 1.119  & 0.656  & 0.549  & 0.481  & 0.509  & 0.436  & 0.438  & 0.397  & 4.405  & 1.538  \\
          & \textbf{48/336} & 4.298  & 1.512  & 1.392  & 0.730  & 0.501  & 0.503  & 1.118  & 0.669  & 0.561  & 0.489  & 0.538  & 0.448  & 0.447  & 0.402  & 4.298  & 1.512  \\
          & \textbf{60/720} & 4.285  & 1.491  & 1.433  & 0.739  & 0.547  & 0.536  & 1.159  & 0.688  & 0.577  & 0.495  & 0.572  & 0.471  & 0.462  & 0.414  & 4.285  & 1.491  \\
          & \textbf{avg} & 4.394  & 1.535  & 1.319  & 0.722  & 0.487  & 0.494  & 1.128  & 0.665  & 0.549  & 0.480  & 0.525  & 0.444  & 0.439  & 0.400  & 4.394  & 1.535  \\
    \bottomrule
    \end{tabular}%
    }
  \label{tbl:exp03_forecast_objective}%
\end{table*}%

\subsubsection{Full Table of Normalization Layer}
\par This section includes the full table of experiments on the normalization layers, as an extended version of Table \ref{tbl:exp04_norm_layer_smaller} in Section \ref{subsubsec:exp_normalization_layer}. 
Results are reported in Table \ref{tbl:exp04_norm_layer}.

\begin{table*}[htbp]
  \centering
  \caption{Full table on the normalization layer, including LayerNorm (LN) and BatchNorm (BN).}
  \vspace{-0.7em}
    \resizebox{0.90\linewidth}{!}{
    \begin{tabular}{c|c|c|cc|cc|cc|cc|cc|cc|cc|cc}
    \toprule
    \multicolumn{3}{c|}{\textbf{dataset}} & \multicolumn{2}{c|}{\textbf{Illness}} & \multicolumn{2}{c|}{\textbf{ETTh1}} & \multicolumn{2}{c|}{\textbf{ETTh2}} & \multicolumn{2}{c|}{\textbf{ETTm1}} & \multicolumn{2}{c|}{\textbf{ETTm2}} & \multicolumn{2}{c|}{\textbf{Weather}} & \multicolumn{2}{c|}{\textbf{ECL}} & \multicolumn{2}{c}{\textbf{Traffic}} \\
    \midrule
    \multicolumn{3}{c|}{\textbf{anomaly sample ratio}} & \multicolumn{2}{c|}{\textbf{0.322}} & \multicolumn{2}{c|}{\textbf{0.127}} & \multicolumn{2}{c|}{\textbf{0.282}} & \multicolumn{2}{c|}{\textbf{0.099}} & \multicolumn{2}{c|}{\textbf{0.160}} & \multicolumn{2}{c|}{\textbf{0.141}} & \multicolumn{2}{c|}{\textbf{0.034}} & \multicolumn{2}{c}{\textbf{0.025}} \\
    \midrule
    \textbf{model} & \textbf{norm} & \textbf{pred\_len} & \textbf{MSE} & \textbf{MAE} & \textbf{MSE} & \textbf{MAE} & \textbf{MSE} & \textbf{MAE} & \textbf{MSE} & \textbf{MAE} & \textbf{MSE} & \textbf{MAE} & \textbf{MSE} & \textbf{MAE} & \textbf{MSE} & \textbf{MAE} & \textbf{MSE} & \textbf{MAE} \\
    \midrule
    \midrule
    \multirow{10}[4]{*}{\textbf{Encoder-only}} & \multirow{5}[2]{*}{\textbf{+LN}} & \textbf{24/96} & 1.261  & 0.764  & \textbf{0.365 } & \textbf{0.401 } & 0.295  & 0.347  & \textbf{0.290 } & \textbf{0.335 } & 0.170  & 0.265  & 0.146  & 0.198  & 0.130  & 0.225  & \textbf{0.350 } & \textbf{0.240 } \\
          &       & \textbf{36/192} & 1.238  & 0.771  & \textbf{0.410 } & \textbf{0.425 } & 0.345  & 0.380  & \textbf{0.330 } & \textbf{0.362 } & 0.240  & 0.308  & \textbf{0.193 } & \textbf{0.243 } & \textbf{0.145 } & \textbf{0.240 } & \textbf{0.374 } & \textbf{0.247 } \\
          &       & \textbf{48/336} & 1.597  & 0.885  & \textbf{0.415 } & \textbf{0.433 } & 0.350  & 0.385  & 0.364  & 0.376  & 0.285  & 0.337  & \textbf{0.245 } & 0.283  & 0.163  & 0.259  & \textbf{0.388 } & \textbf{0.264 } \\
          &       & \textbf{60/720} & 1.621  & 0.899  & \textbf{0.442 } & \textbf{0.465 } & 0.395  & 0.428  & \textbf{0.405 } & \textbf{0.410 } & 0.379  & 0.400  & \textbf{0.323 } & \textbf{0.337 } & \textbf{0.193 } & \textbf{0.284 } & \textbf{0.415 } & \textbf{0.282 } \\
          &       & \textbf{avg} & 1.429  & 0.830  & \textbf{0.408 } & \textbf{0.431 } & 0.346  & 0.385  & \textbf{0.347 } & \textbf{0.371 } & 0.269  & 0.328  & \textbf{0.227 } & \textbf{0.265 } & 0.158  & 0.252  & \textbf{0.382 } & \textbf{0.258 } \\
\cmidrule{2-19}          & \multirow{5}[2]{*}{\textbf{+BN}} & \textbf{24/96} & \textbf{1.156 } & \textbf{0.697 } & 0.367  & 0.402  & \textbf{0.275 } & \textbf{0.336 } & 0.294  & 0.338  & \textbf{0.163 } & \textbf{0.261 } & \textbf{0.145 } & \textbf{0.197 } & \textbf{0.128 } & \textbf{0.220 } & \textbf{0.350 } & \textbf{0.240 } \\
          &       & \textbf{36/192} & \textbf{1.096 } & \textbf{0.691 } & 0.413  & 0.428  & \textbf{0.337 } & \textbf{0.375 } & 0.336  & 0.368  & \textbf{0.223 } & \textbf{0.301 } & \textbf{0.193 } & \textbf{0.243 } & \textbf{0.145 } & \textbf{0.240 } & \textbf{0.374 } & \textbf{0.247 } \\
          &       & \textbf{48/336} & \textbf{1.407 } & \textbf{0.802 } & 0.419  & 0.437  & \textbf{0.343 } & \textbf{0.379 } & \textbf{0.363 } & \textbf{0.375 } & \textbf{0.279 } & \textbf{0.335 } & \textbf{0.245 } & \textbf{0.282 } & \textbf{0.162 } & \textbf{0.256 } & 0.390  & 0.265  \\
          &       & \textbf{60/720} & \textbf{1.453 } & \textbf{0.771 } & \textbf{0.442 } & \textbf{0.465 } & \textbf{0.383 } & \textbf{0.418 } & 0.410  & 0.415  & \textbf{0.375 } & \textbf{0.397 } & 0.324  & 0.338  & 0.194  & 0.285  & 0.417  & 0.284  \\
          &       & \textbf{avg} & \textbf{1.278 } & \textbf{0.740 } & 0.410  & 0.433  & \textbf{0.335 } & \textbf{0.377 } & 0.351  & 0.374  & \textbf{0.260 } & \textbf{0.324 } & \textbf{0.227 } & \textbf{0.265 } & \textbf{0.157 } & \textbf{0.250 } & 0.383  & 0.259  \\
    \midrule
    \midrule
    \multirow{10}[2]{*}{\textbf{Encoder-decoder}} & \multirow{5}[1]{*}{\textbf{+LN}} & \textbf{24/96} & 1.382  & 0.779  & \textbf{0.503 } & \textbf{0.448 } & \textbf{0.284 } & \textbf{0.347 } & \textbf{0.313 } & \textbf{0.342 } & 0.198  & 0.274  & \textbf{0.142 } & \textbf{0.191 } & \textbf{0.128 } & \textbf{0.220 } & \textbf{0.349 } & \textbf{0.239 } \\
          &       & \textbf{36/192} & 1.279  & 0.735  & \textbf{0.502 } & \textbf{0.449 } & \textbf{0.374 } & \textbf{0.399 } & \textbf{0.373 } & \textbf{0.390 } & \textbf{0.267 } & \textbf{0.309 } & 0.198  & 0.251  & \textbf{0.145 } & \textbf{0.236 } & \textbf{0.376 } & \textbf{0.242 } \\
          &       & \textbf{48/336} & 1.398  & 0.798  & 0.495  & 0.462  & 0.352  & 0.429  & \textbf{0.375 } & 0.412  & 0.318  & 0.367  & 0.265  & 0.304  & 0.166  & 0.255  & \textbf{0.398 } & \textbf{0.274 } \\
          &       & \textbf{60/720} & 1.702  & 0.859  & 0.539  & 0.486  & 0.418  & 0.461  & \textbf{0.524 } & 0.480  & 0.394  & 0.416  & 0.325  & 0.348  & \textbf{0.198 } & \textbf{0.289 } & \textbf{0.422 } & \textbf{0.292 } \\
          &       & \textbf{avg} & 1.440  & 0.793  & \textbf{0.510 } & \textbf{0.461 } & \textbf{0.357 } & 0.409  & \textbf{0.396 } & 0.406  & 0.294  & 0.342  & 0.233  & 0.274  & \textbf{0.159 } & \textbf{0.250 } & \textbf{0.386 } & \textbf{0.262 } \\
\cmidrule{2-19}          & \multirow{5}[1]{*}{\textbf{+BN}} & \textbf{24/96} & \textbf{1.301 } & \textbf{0.755 } & 0.506  & 0.454  & 0.299  & 0.352  & \textbf{0.313 } & \textbf{0.342 } & \textbf{0.187 } & \textbf{0.271 } & 0.144  & 0.193  & \textbf{0.128 } & 0.221  & 0.351  & 0.242  \\
          &       & \textbf{36/192} & \textbf{1.154 } & \textbf{0.710 } & 0.509  & 0.455  & 0.384  & 0.405  & 0.377  & 0.396  & 0.274  & 0.320  & \textbf{0.196 } & \textbf{0.250 } & 0.147  & 0.239  & 0.378  & 0.248  \\
          &       & \textbf{48/336} & \textbf{1.320 } & \textbf{0.771 } & \textbf{0.490 } & \textbf{0.467 } & \textbf{0.344 } & \textbf{0.418 } & 0.381  & \textbf{0.409 } & \textbf{0.309 } & \textbf{0.363 } & \textbf{0.259 } & \textbf{0.298 } & \textbf{0.165 } & \textbf{0.257 } & \textbf{0.398 } & \textbf{0.274 } \\
          &       & \textbf{60/720} & \textbf{1.558 } & \textbf{0.821 } & \textbf{0.533 } & \textbf{0.476 } & \textbf{0.410 } & \textbf{0.448 } & 0.536  & \textbf{0.473 } & \textbf{0.386 } & \textbf{0.410 } & \textbf{0.318 } & \textbf{0.340 } & \textbf{0.198 } & 0.291  & \textbf{0.422 } & \textbf{0.292 } \\
          &       & \textbf{avg} & \textbf{1.333 } & \textbf{0.764 } & \textbf{0.510 } & 0.463  & 0.359  & \textbf{0.406 } & 0.402  & \textbf{0.405 } & \textbf{0.289 } & \textbf{0.341 } & \textbf{0.229 } & \textbf{0.270 } & 0.160  & 0.252  & 0.387  & 0.264  \\
    \midrule
    \midrule
    \multirow{10}[3]{*}{\textbf{Decoder-only}} & \multirow{5}[1]{*}{\textbf{+LN}} & \textbf{24/96} & \textbf{1.705 } & \textbf{0.888 } & \textbf{0.591 } & \textbf{0.509 } & 0.399  & 0.429  & \textbf{0.310 } & \textbf{0.341 } & 0.258  & 0.314  & 0.156  & 0.211  & 0.129  & 0.222  & \textbf{0.348 } & \textbf{0.242 } \\
          &       & \textbf{36/192} & \textbf{1.903 } & \textbf{0.939 } & \textbf{0.540 } & \textbf{0.470 } & 0.425  & \textbf{0.436 } & \textbf{0.333 } & \textbf{0.369 } & 0.275  & 0.349  & 0.217  & 0.267  & 0.149  & 0.244  & \textbf{0.371 } & \textbf{0.249 } \\
          &       & \textbf{48/336} & \textbf{1.473 } & \textbf{0.812 } & 0.455  & 0.439  & 0.497  & 0.504  & \textbf{0.374 } & \textbf{0.407 } & 0.393  & 0.416  & 0.288  & 0.312  & \textbf{0.162 } & \textbf{0.258 } & 0.398  & 0.273  \\
          &       & \textbf{60/720} & \textbf{1.997 } & \textbf{0.972 } & 0.625  & 0.597  & 0.581  & 0.539  & \textbf{0.548 } & \textbf{0.471 } & 0.417  & 0.434  & 0.358  & 0.364  & 0.201  & 0.293  & 0.424  & 0.292  \\
          &       & \textbf{avg} & \textbf{1.770 } & \textbf{0.903 } & 0.553  & \textbf{0.504 } & 0.476  & 0.477  & \textbf{0.391 } & \textbf{0.397 } & 0.336  & 0.378  & 0.255  & 0.289  & 0.160  & 0.254  & 0.385  & \textbf{0.264 } \\
\cmidrule{2-19}          & \multirow{5}[2]{*}{\textbf{+BN}} & \textbf{24/96} & 2.124  & 0.956  & 0.594  & 0.515  & \textbf{0.390 } & \textbf{0.425 } & \textbf{0.310 } & \textbf{0.341 } & \textbf{0.244 } & \textbf{0.307 } & \textbf{0.152 } & \textbf{0.205 } & \textbf{0.128 } & \textbf{0.221 } & 0.350  & 0.244  \\
          &       & \textbf{36/192} & 2.241  & 0.987  & 0.547  & 0.476  & \textbf{0.417 } & 0.445  & 0.341  & 0.371  & \textbf{0.273 } & \textbf{0.343 } & \textbf{0.207 } & \textbf{0.260 } & \textbf{0.147 } & \textbf{0.242 } & \textbf{0.371 } & \textbf{0.249 } \\
          &       & \textbf{48/336} & 1.943  & 0.932  & \textbf{0.452 } & \textbf{0.441 } & \textbf{0.437 } & \textbf{0.481 } & 0.391  & 0.407  & \textbf{0.366 } & \textbf{0.403 } & \textbf{0.273 } & \textbf{0.303 } & 0.163  & \textbf{0.258 } & \textbf{0.396 } & \textbf{0.272 } \\
          &       & \textbf{60/720} & 2.187  & 0.981  & \textbf{0.615 } & \textbf{0.586 } & \textbf{0.520 } & \textbf{0.504 } & 0.560  & 0.476  & \textbf{0.413 } & \textbf{0.427 } & \textbf{0.349 } & \textbf{0.356 } & \textbf{0.199 } & \textbf{0.290 } & \textbf{0.420 } & \textbf{0.290 } \\
          &       & \textbf{avg} & 2.124  & 0.964  & \textbf{0.552 } & 0.505  & \textbf{0.441 } & \textbf{0.464 } & 0.401  & 0.399  & \textbf{0.324 } & \textbf{0.370 } & \textbf{0.245 } & \textbf{0.281 } & \textbf{0.159 } & \textbf{0.253 } & \textbf{0.384 } & \textbf{0.264 } \\
    \bottomrule
    \end{tabular}%
    }
  \label{tbl:exp04_norm_layer}%
\end{table*}%

\subsubsection{Full Table of Variable Forecasting Length}
\par This section includes the full table of experiments on the variable forecasting length setting, as an extended version of Table \ref{tbl:exp05_multi_pred_len_smaller} in Section \ref{subsubsec:exp_forecasting_length_setting}. 
Results are reported in Table \ref{tbl:exp05_multi_pred_len}.

\begin{table*}[htbp]
  \centering
  \caption{Full table on variable forecasting length setting. For simplicity, we use these abbreviations in model names: ``aggr." for ``aggregation”, ``direct." for ``direct-mapping", and ``autoreg." for the ``autoregressive" paradigm.}
  \vspace{-0.7em}
    \resizebox{0.90\linewidth}{!}{
    \begin{tabular}{c|c|c|cc|cc|cc|cc|cc|cc|cc|cc}
    \toprule
    \multicolumn{3}{c|}{\textbf{dataset}} & \multicolumn{2}{c|}{\textbf{Illness}} & \multicolumn{2}{c|}{\textbf{ETTh1}} & \multicolumn{2}{c|}{\textbf{ETTh2}} & \multicolumn{2}{c|}{\textbf{ETTm1}} & \multicolumn{2}{c|}{\textbf{ETTm2}} & \multicolumn{2}{c|}{\textbf{Weather}} & \multicolumn{2}{c|}{\textbf{ECL}} & \multicolumn{2}{c}{\textbf{Traffic}} \\
    \midrule
    \textbf{No.} & \textbf{model} & \textbf{pred\_len} & \textbf{MSE} & \textbf{MAE} & \textbf{MSE} & \textbf{MAE} & \textbf{MSE} & \textbf{MAE} & \textbf{MSE} & \textbf{MAE} & \textbf{MSE} & \textbf{MAE} & \textbf{MSE} & \textbf{MAE} & \textbf{MSE} & \textbf{MAE} & \textbf{MSE} & \textbf{MAE} \\
    \midrule
    \midrule
    \multirow{5}[1]{*}{1} & \multirow{5}[1]{*}{\makecell{\textbf{Encoder-only +} \\ \textbf{complete aggr. + direct.}}} & \textbf{24/96} & \textbf{1.325 } & \textbf{0.772 } & \textbf{0.387 } & \textbf{0.417 } & \textbf{0.308 } & \textbf{0.358 } & \textbf{0.318 } & \textbf{0.365 } & \textbf{0.177 } & \textbf{0.266 } & \textbf{0.154 } & \textbf{0.210 } & 0.135  & 0.235  & 0.372  & 0.267  \\
          & & \textbf{36/192} & \textbf{1.295 } & \textbf{0.766 } & \textbf{0.420 } & \textbf{0.436 } & \textbf{0.366 } & \textbf{0.395 } & \textbf{0.350 } & \textbf{0.383 } & \textbf{0.229 } & \textbf{0.302 } & \textbf{0.198 } & \textbf{0.248 } & 0.150  & 0.248  & 0.386  & 0.273  \\
          & & \textbf{48/336} & \textbf{1.307 } & \textbf{0.769 } & \textbf{0.425 } & \textbf{0.441 } & \textbf{0.372 } & \textbf{0.409 } & \textbf{0.375 } & \textbf{0.399 } & \textbf{0.280 } & \textbf{0.336 } & \textbf{0.247 } & \textbf{0.285 } & 0.166  & 0.264  & 0.397  & 0.279  \\
          & & \textbf{60/720} & \textbf{1.453 } & \textbf{0.771 } & \textbf{0.442 } & \textbf{0.465 } & \textbf{0.383 } & \textbf{0.418 } & \textbf{0.410 } & \textbf{0.415 } & \textbf{0.375 } & \textbf{0.397 } & 0.324  & \textbf{0.338 } & \textbf{0.194 } & \textbf{0.285 } & \textbf{0.417 } & \textbf{0.284 } \\
          & & \textbf{avg} & \textbf{1.345 } & \textbf{0.770 } & \textbf{0.419 } & \textbf{0.440 } & \textbf{0.357 } & \textbf{0.395 } & \textbf{0.363 } & \textbf{0.391 } & \textbf{0.265 } & \textbf{0.325 } & \textbf{0.231 } & \textbf{0.270 } & 0.161  & 0.258  & 0.393  & 0.276  \\
    \midrule
    \multirow{5}[1]{*}{2} & \multirow{5}[1]{*}{\makecell{\textbf{Encoder-only +} \\ \textbf{no aggr. + direct.}}} & \textbf{24/96} & 1.412  & 0.779  & 0.435  & 0.431  & 0.368  & 0.400  & 0.336  & 0.374  & 0.236  & 0.324  & 0.171  & 0.228  & \textbf{0.131 } & 0.227  & 0.359  & 0.250  \\
          & & \textbf{36/192} & 1.396  & 0.776  & 0.498  & 0.443  & 0.425  & 0.440  & 0.363  & 0.390  & 0.278  & 0.347  & 0.212  & 0.262  & \textbf{0.147 } & 0.242  & \textbf{0.376 } & 0.257  \\
          & & \textbf{48/336} & 1.373  & 0.780  & 0.458  & 0.436  & 0.439  & 0.459  & 0.393  & 0.409  & 0.319  & 0.372  & 0.259  & 0.296  & \textbf{0.163 } & \textbf{0.257 } & \textbf{0.387 } & \textbf{0.262 } \\
          & & \textbf{60/720} & 1.534  & 0.830  & 0.525  & 0.502  & 0.493  & 0.496  & 0.415  & 0.421  & 0.390  & 0.412  & 0.345  & 0.349  & 0.197  & 0.287  & 0.420  & 0.290  \\
          & & \textbf{avg} & 1.429  & 0.791  & 0.479  & 0.453  & 0.431  & 0.449  & 0.377  & 0.399  & 0.306  & 0.364  & 0.247  & 0.284  & \textbf{0.160 } & \textbf{0.253 } & \textbf{0.386 } & 0.265  \\
    \midrule
    \multirow{5}[1]{*}{3} & \multirow{5}[1]{*}{\makecell{\textbf{Prefix decoder +} \\ \textbf{no aggr. + direct.}}} & \textbf{24/96} & 1.492  & 0.798  & 0.451  & 0.430  & 0.356  & 0.402  & 0.387  & 0.419  & 0.239  & 0.334  & 0.179  & 0.246  & 0.134  & 0.232  & \textbf{0.357 } & 0.247  \\
          & & \textbf{36/192} & 1.478  & 0.809  & 0.505  & 0.461  & 0.405  & 0.433  & 0.433  & 0.440  & 0.290  & 0.363  & 0.220  & 0.279  & 0.152  & 0.250  & \textbf{0.376 } & 0.255  \\
          & & \textbf{48/336} & 1.336  & 0.776  & 0.499  & 0.456  & 0.442  & 0.474  & 0.480  & 0.461  & 0.336  & 0.391  & 0.262  & 0.309  & 0.171  & 0.269  & 0.394  & 0.262  \\
          & & \textbf{60/720} & 1.558  & 0.821  & 0.563  & 0.530  & 0.456  & 0.481  & 0.519  & 0.458  & 0.413  & 0.431  & 0.328  & 0.349  & 0.197  & 0.290  & 0.418  & 0.288  \\
          & & \textbf{avg} & 1.466  & 0.801  & 0.505  & 0.469  & 0.415  & 0.448  & 0.455  & 0.445  & 0.320  & 0.380  & 0.247  & 0.296  & 0.164  & 0.260  & \textbf{0.386 } & 0.263  \\
    \midrule
    \multirow{5}[1]{*}{4} & \multirow{5}[0]{*}{\makecell{\textbf{Encoder-decoder +} \\ \textbf{no aggr. + direct.}}} & \textbf{24/96} & 1.879  & 0.816  & 0.516  & 0.460  & 0.374  & 0.413  & 0.337  & 0.376  & 0.229  & 0.321  & 0.166  & 0.223  & 0.132  & \textbf{0.225 } & 0.359  & 0.248  \\
          & & \textbf{36/192} & 1.797  & 0.803  & 0.518  & 0.462  & 0.391  & 0.425  & 0.381  & 0.400  & 0.274  & 0.345  & 0.209  & 0.260  & 0.150  & \textbf{0.241 } & 0.378  & 0.248  \\
          & & \textbf{48/336} & 1.740  & 0.805  & 0.505  & 0.478  & 0.401  & 0.438  & 0.419  & 0.423  & 0.317  & 0.371  & 0.257  & 0.296  & 0.167  & 0.258  & 0.395  & 0.270  \\
          & & \textbf{60/720} & 1.935  & 0.843  & 0.533  & 0.476  & 0.410  & 0.448  & 0.536  & 0.473  & 0.387  & 0.407  & \textbf{0.318 } & 0.340  & 0.198  & 0.291  & 0.422  & 0.292  \\
          & & \textbf{avg} & 1.838  & 0.817  & 0.518  & 0.469  & 0.394  & 0.431  & 0.418  & 0.418  & 0.302  & 0.361  & 0.238  & 0.280  & 0.162  & 0.254  & 0.389  & 0.265  \\
    \midrule
    \multirow{5}[1]{*}{5} & \multirow{5}[1]{*}{\makecell{\textbf{Decoder-only +} \\ \textbf{no aggr. + direct.}}} & \textbf{24/96} & 2.071  & 0.936  & 0.463  & 0.451  & 0.406  & 0.421  & 0.360  & 0.385  & 0.259  & 0.340  & 0.185  & 0.253  & 0.132  & 0.230  & \textbf{0.357 } & \textbf{0.246 } \\
          & & \textbf{36/192} & 2.176  & 0.962  & 0.552  & 0.480  & 0.489  & 0.479  & 0.397  & 0.406  & 0.306  & 0.367  & 0.225  & 0.285  & 0.151  & 0.248  & \textbf{0.376 } & \textbf{0.253 } \\
          & & \textbf{48/336} & 2.054  & 0.935  & 0.596  & 0.517  & 0.507  & 0.498  & 0.427  & 0.425  & 0.347  & 0.392  & 0.266  & 0.315  & 0.170  & 0.268  & 0.392  & 0.260  \\
          & & \textbf{60/720} & 2.187  & 0.981  & 0.615  & 0.586  & 0.520  & 0.504  & 0.560  & 0.476  & 0.413  & 0.427  & 0.349  & 0.356  & 0.199  & 0.290  & 0.420  & 0.290  \\
          & & \textbf{avg} & 2.122  & 0.954  & 0.557  & 0.509  & 0.481  & 0.476  & 0.436  & 0.423  & 0.331  & 0.382  & 0.256  & 0.302  & 0.163  & 0.259  & \textbf{0.386 } & \textbf{0.262 } \\
    \midrule
    \multirow{5}[1]{*}{6} & \multirow{5}[2]{*}{\makecell{\textbf{Encoder-decoder +} \\ \textbf{no aggr. + autoerg.}}} & \textbf{24/96} & 4.589  & 1.598  & 1.150  & 0.698  & 0.418  & 0.451  & 1.115  & 0.645  & 0.509  & 0.455  & 0.481  & 0.420  & 0.408  & 0.386  & 4.589  & 1.598  \\
          & & \textbf{36/192} & 4.405  & 1.538  & 1.301  & 0.721  & 0.482  & 0.487  & 1.119  & 0.656  & 0.549  & 0.481  & 0.509  & 0.436  & 0.438  & 0.397  & 4.405  & 1.538  \\
          & & \textbf{48/336} & 4.298  & 1.512  & 1.392  & 0.730  & 0.501  & 0.503  & 1.118  & 0.669  & 0.561  & 0.489  & 0.538  & 0.448  & 0.447  & 0.402  & 4.298  & 1.512  \\
          & & \textbf{60/720} & 4.285  & 1.491  & 1.433  & 0.739  & 0.547  & 0.536  & 1.159  & 0.688  & 0.577  & 0.495  & 0.572  & 0.471  & 0.462  & 0.414  & 4.285  & 1.491  \\
          & & \textbf{avg} & 4.394  & 1.535  & 1.319  & 0.722  & 0.487  & 0.494  & 1.128  & 0.665  & 0.549  & 0.480  & 0.525  & 0.444  & 0.439  & 0.400  & 4.394  & 1.535  \\
    \midrule
    \multirow{5}[1]{*}{7} & \multirow{5}[2]{*}{\makecell{\textbf{Decoder-only +} \\ \textbf{no aggr. + autoreg.}}} & \textbf{24/96} & 5.795  & 1.780  & 1.012  & 0.749  & 0.604  & 0.527  & 1.321  & 0.735  & 0.542  & 0.483  & 0.502  & 0.443  & 0.432  & 0.398  & 5.795  & 1.780  \\
          & & \textbf{36/192} & 5.156  & 1.615  & 1.234  & 0.760  & 0.612  & 0.531  & 1.344  & 0.742  & 0.601  & 0.512  & 0.531  & 0.451  & 0.461  & 0.408  & 5.156  & 1.615  \\
          & & \textbf{48/336} & 5.070  & 1.588  & 1.338  & 0.768  & 0.663  & 0.558  & 1.378  & 0.760  & 0.621  & 0.521  & 0.555  & 0.462  & 0.471  & 0.412  & 5.070  & 1.588  \\
          & & \textbf{60/720} & 5.251  & 1.624  & 1.349  & 0.787  & 0.608  & 0.545  & 1.363  & 0.759  & 0.629  & 0.527  & 0.584  & 0.489  & 0.494  & 0.421  & 5.251  & 1.624  \\
          & & \textbf{avg} & 5.318  & 1.652  & 1.233  & 0.766  & 0.622  & 0.540  & 1.352  & 0.749  & 0.598  & 0.511  & 0.543  & 0.461  & 0.465  & 0.410  & 5.318  & 1.652  \\
    \bottomrule
    \end{tabular}%
    }
  \label{tbl:exp05_multi_pred_len}%
\end{table*}%

\subsubsection{Full Table of Combined Model}
\par This section includes the full table of experiments on the combined optimal model compared to other existing models, as an extended version of Table \ref{tbl:exp06_combination_model_smaller} in Section \ref{subsec:combination_model}. 
Results are reported in Table \ref{tbl:exp06_combination_model}.

\begin{table*}[htbp]
  \centering
  \caption{Full table on the comparison between the combined optimal model and several existing LTSF models.}
  \vspace{-0.7em}
  \resizebox{0.92\linewidth}{!}{
    \begin{tabular}{c|c|cc|cc|cc|cc|cc|cc|cc|cc|c}
    \toprule
    \multicolumn{2}{c|}{\textbf{dataset}} & \multicolumn{2}{c|}{\textbf{Illness}} & \multicolumn{2}{c|}{\textbf{ETTh1}} & \multicolumn{2}{c|}{\textbf{ETTh2}} & \multicolumn{2}{c|}{\textbf{ETTm1}} & \multicolumn{2}{c|}{\textbf{ETTm2}} & \multicolumn{2}{c|}{\textbf{Weather}} & \multicolumn{2}{c|}{\textbf{ECL}} & \multicolumn{2}{c|}{\textbf{Traffic}} & \multirow{2}[4]{*}{\textbf{1st Count}} \\
\cmidrule{1-18}    \textbf{model name} & \textbf{pred\_len} & \textbf{MSE} & \textbf{MAE} & \textbf{MSE} & \textbf{MAE} & \textbf{MSE} & \textbf{MAE} & \textbf{MSE} & \textbf{MAE} & \textbf{MSE} & \textbf{MAE} & \textbf{MSE} & \textbf{MAE} & \textbf{MSE} & \textbf{MAE} & \textbf{MSE} & \textbf{MAE} &  \\
    \midrule
    \midrule
    \multirow{5}[2]{*}{\textbf{Combined Model}} & \textbf{24/96} & \textbf{1.156 } & \textbf{0.697 } & 0.367  & 0.402  & 0.275  & \textbf{0.336 } & 0.294  & \textbf{0.338 } & 0.163  & 0.261  & \textbf{0.145 } & \textbf{0.197 } & \textbf{0.128 } & \textbf{0.220 } & \textbf{0.350 } & \textbf{0.240 } & \multirow{5}[2]{*}{47} \\
          & \textbf{36/192} & \textbf{1.096 } & \textbf{0.691 } & 0.413  & 0.428  & 0.337  & 0.375  & 0.336  & \textbf{0.368 } & \textbf{0.223 } & 0.301  & \textbf{0.193 } & 0.243  & \textbf{0.145 } & \textbf{0.240 } & \textbf{0.374 } & \textbf{0.247 } &  \\
          & \textbf{48/336} & \textbf{1.407 } & \textbf{0.802 } & \textbf{0.419 } & \textbf{0.437 } & 0.343  & \textbf{0.379 } & \textbf{0.363 } & \textbf{0.375 } & 0.279  & 0.335  & \textbf{0.245 } & \textbf{0.282 } & \textbf{0.162 } & \textbf{0.256 } & 0.390  & 0.265  &  \\
          & \textbf{60/720} & \textbf{1.453 } & \textbf{0.771 } & 0.442  & 0.465  & 0.383  & \textbf{0.418 } & 0.410  & \textbf{0.415 } & 0.375  & 0.397  & 0.324  & 0.338  & 0.194  & \textbf{0.285 } & 0.417  & \textbf{0.284 } &  \\
          & \textbf{avg} & \textbf{1.278 } & \textbf{0.740 } & \textbf{0.410 } & \textbf{0.433 } & 0.335  & \textbf{0.377 } & 0.351  & \textbf{0.374 } & 0.260  & 0.324  & \textbf{0.227 } & \textbf{0.265 } & \textbf{0.157 } & \textbf{0.250 } & \textbf{0.383 } & \textbf{0.259 } &  \\
    \midrule
    \multirow{5}[1]{*}{\textbf{FEDformer}} & \textbf{24/96} & 2.624  & 1.095  & 0.376  & 0.415  & 0.332  & 0.374  & 0.326  & 0.390  & 0.180  & 0.271  & 0.238  & 0.314  & 0.186  & 0.302  & 0.576  & 0.359  & \multirow{5}[1]{*}{0} \\
          & \textbf{36/192} & 2.516  & 1.021  & 0.423  & 0.446  & 0.407  & 0.446  & 0.365  & 0.415  & 0.252  & 0.318  & 0.275  & 0.329  & 0.197  & 0.311  & 0.610  & 0.380  &  \\
          & \textbf{48/336} & 2.505  & 1.041  & 0.444  & 0.462  & 0.400  & 0.447  & 0.392  & 0.425  & 0.324  & 0.364  & 0.339  & 0.377  & 0.213  & 0.328  & 0.608  & 0.375  &  \\
          & \textbf{60/720} & 2.742  & 1.122  & 0.469  & 0.492  & 0.412  & 0.469  & 0.446  & 0.458  & 0.362  & 0.385  & 0.389  & 0.409  & 0.233  & 0.344  & 0.621  & 0.375  &  \\
          & \textbf{avg} & 2.597  & 1.070  & 0.428  & 0.454  & 0.388  & 0.434  & 0.382  & 0.422  & 0.410  & 0.420  & 0.310  & 0.357  & 0.207  & 0.321  & 0.604  & 0.372  &  \\
    \midrule
    \multirow{5}[0]{*}{\textbf{PatchTST}} & \textbf{24/96} & 1.319  & 0.754  & 0.370  & 0.400  & 0.274  & 0.337  & 0.295  & 0.346  & 0.166  & 0.256  & 0.149  & 0.198  & 0.129  & 0.222  & 0.360  & 0.249  & \multirow{5}[0]{*}{11} \\
          & \textbf{36/192} & 1.579  & 0.870  & 0.414  & 0.429  & 0.341  & 0.382  & 0.333  & 0.370  & 0.224  & \textbf{0.296 } & 0.194  & \textbf{0.241 } & 0.147  & \textbf{0.240 } & 0.379  & 0.256  &  \\
          & \textbf{48/336} & 1.553  & 0.815  & 0.422  & 0.440  & 0.329  & 0.384  & 0.369  & 0.392  & 0.274  & \textbf{0.329 } & 0.246  & 0.284  & 0.163  & 0.259  & 0.392  & 0.266  &  \\
          & \textbf{60/720} & 1.470  & 0.788  & 0.447  & 0.468  & \textbf{0.379 } & 0.422  & 0.416  & 0.420  & \textbf{0.362 } & \textbf{0.385 } & 0.318  & 0.336  & 0.197  & 0.290  & 0.432  & 0.286  &  \\
          & \textbf{avg} & 1.480  & 0.807  & 0.413  & 0.434  & 0.331  & 0.381  & 0.353  & 0.382  & \textbf{0.257 } & \textbf{0.317 } & \textbf{0.227 } & \textbf{0.265 } & 0.159  & 0.253  & 0.391  & 0.264  &  \\
    \midrule
    \multirow{5}[1]{*}{\textbf{iTransformer}} & \textbf{24/96} & 2.361  & 1.053  & 0.400  & 0.425  & 0.299  & 0.359  & 0.311  & 0.366  & 0.179  & 0.273  & 0.168  & 0.220  & 0.133  & 0.229  & 0.354  & 0.259  & \multirow{5}[1]{*}{3} \\
          & \textbf{36/192} & 2.243  & 1.011  & 0.427  & 0.443  & 0.377  & 0.406  & 0.348  & 0.385  & 0.242  & 0.315  & 0.209  & 0.254  & 0.152  & 0.248  & 0.379  & 0.268  &  \\
          & \textbf{48/336} & 2.148  & 1.004  & 0.457  & 0.465  & 0.429  & 0.442  & 0.379  & 0.405  & 0.291  & 0.345  & 0.266  & 0.295  & 0.169  & 0.265  & 0.394  & 0.273  &  \\
          & \textbf{60/720} & 2.069  & 0.991  & 0.631  & 0.574  & 0.444  & 0.466  & 0.443  & 0.444  & 0.377  & 0.398  & 0.341  & 0.345  & \textbf{0.192 } & \textbf{0.285 } & \textbf{0.416 } & 0.288  &  \\
          & \textbf{avg} & 2.205  & 1.015  & 0.479  & 0.477  & 0.387  & 0.418  & 0.370  & 0.400  & 0.272  & 0.333  & 0.246  & 0.279  & 0.162  & 0.257  & 0.386  & 0.272  &  \\
    \midrule
    \multirow{5}[1]{*}{\textbf{CATS}} & \textbf{24/96} & 2.729  & 1.087  & 0.386  & 0.409  & \textbf{0.268 } & \textbf{0.336 } & \textbf{0.283 } & 0.340  & 0.169  & 0.263  & 0.146  & 0.199  & 0.129  & 0.221  & 0.352  & 0.243  & \multirow{5}[1]{*}{15} \\
          & \textbf{36/192} & 4.324  & 1.499  & 0.419  & 0.431  & \textbf{0.322 } & \textbf{0.365 } & \textbf{0.323 } & 0.369  & 0.224  & 0.305  & 0.194  & 0.244  & 0.147  & 0.241  & 0.375  & 0.253  &  \\
          & \textbf{48/336} & 4.006  & 1.377  & 0.422  & 0.438  & \textbf{0.333 } & 0.382  & 0.368  & 0.389  & \textbf{0.273 } & 0.333  & 0.247  & 0.283  & 0.163  & 0.258  & \textbf{0.387 } & \textbf{0.260 } &  \\
          & \textbf{60/720} & 4.076  & 1.406  & \textbf{0.431 } & \textbf{0.458 } & 0.398  & 0.440  & \textbf{0.403 } & 0.417  & 0.364  & 0.389  & 0.324  & 0.340  & 0.195  & 0.284  & 0.425  & 0.286  &  \\
          & \textbf{avg} & 3.784  & 1.342  & 0.415  & 0.434  & \textbf{0.330 } & 0.381  & \textbf{0.344 } & 0.379  & 0.258  & 0.323  & 0.228  & 0.267  & 0.159  & 0.251  & 0.385  & 0.261  &  \\
    \midrule
    \multirow{5}[0]{*}{\textbf{ARMA-Attention}} & \textbf{24/96} & 1.692  & 0.821  & \textbf{0.361 } & \textbf{0.398 } & 0.271  & 0.337  & 0.301  & 0.350  & \textbf{0.158 } & \textbf{0.251 } & 0.152  & 0.202  & 0.131  & 0.223  & 0.369  & 0.255  & \multirow{5}[0]{*}{14} \\
          & \textbf{36/192} & 2.140  & 1.003  & \textbf{0.397 } & \textbf{0.422 } & 0.337  & 0.377  & 0.333  & 0.369  & \textbf{0.223 } & 0.303  & 0.198  & 0.245  & 0.149  & 0.243  & 0.385  & 0.267  &  \\
          & \textbf{48/336} & 2.132  & 0.987  & \textbf{0.419 } & 0.439  & 0.366  & 0.391  & 0.366  & 0.381  & 0.279  & 0.334  & \textbf{0.245 } & \textbf{0.282 } & 0.165  & 0.261  & 0.400  & 0.269  &  \\
          & \textbf{60/720} & 1.991  & 0.956  & 0.445  & 0.472  & 0.393  & 0.425  & 0.414  & 0.419  & 0.381  & 0.400  & \textbf{0.316 } & \textbf{0.334 } & 0.200  & 0.295  & 0.451  & 0.295  &  \\
          & \textbf{avg} & 1.989  & 0.942  & 0.406  & \textbf{0.433 } & 0.342  & 0.383  & 0.354  & 0.380  & 0.260  & 0.322  & 0.228  & 0.266  & 0.161  & 0.256  & 0.401  & 0.272  &  \\
    \bottomrule
    \end{tabular}%
    }
  \label{tbl:exp06_combination_model}%
\end{table*}%

\section{Attention Map Visualization}
\par
\begin{figure*}[t!]
    \centering  %
    \subfigure[Encoder]{
        \includegraphics[width=0.235\textwidth]{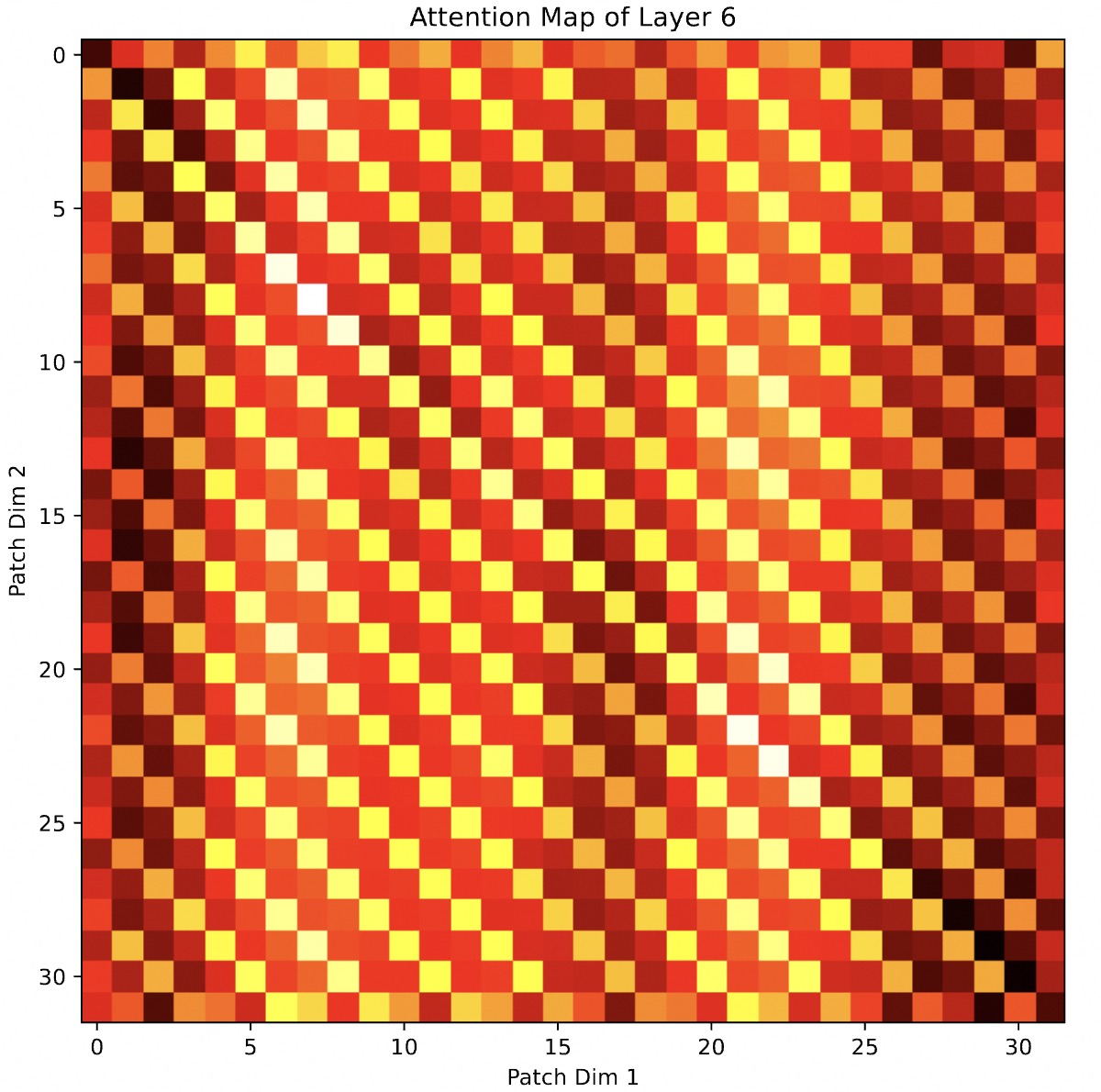}
    }%
    \subfigure[Prefix decoder]{
        \includegraphics[width=0.235\textwidth]{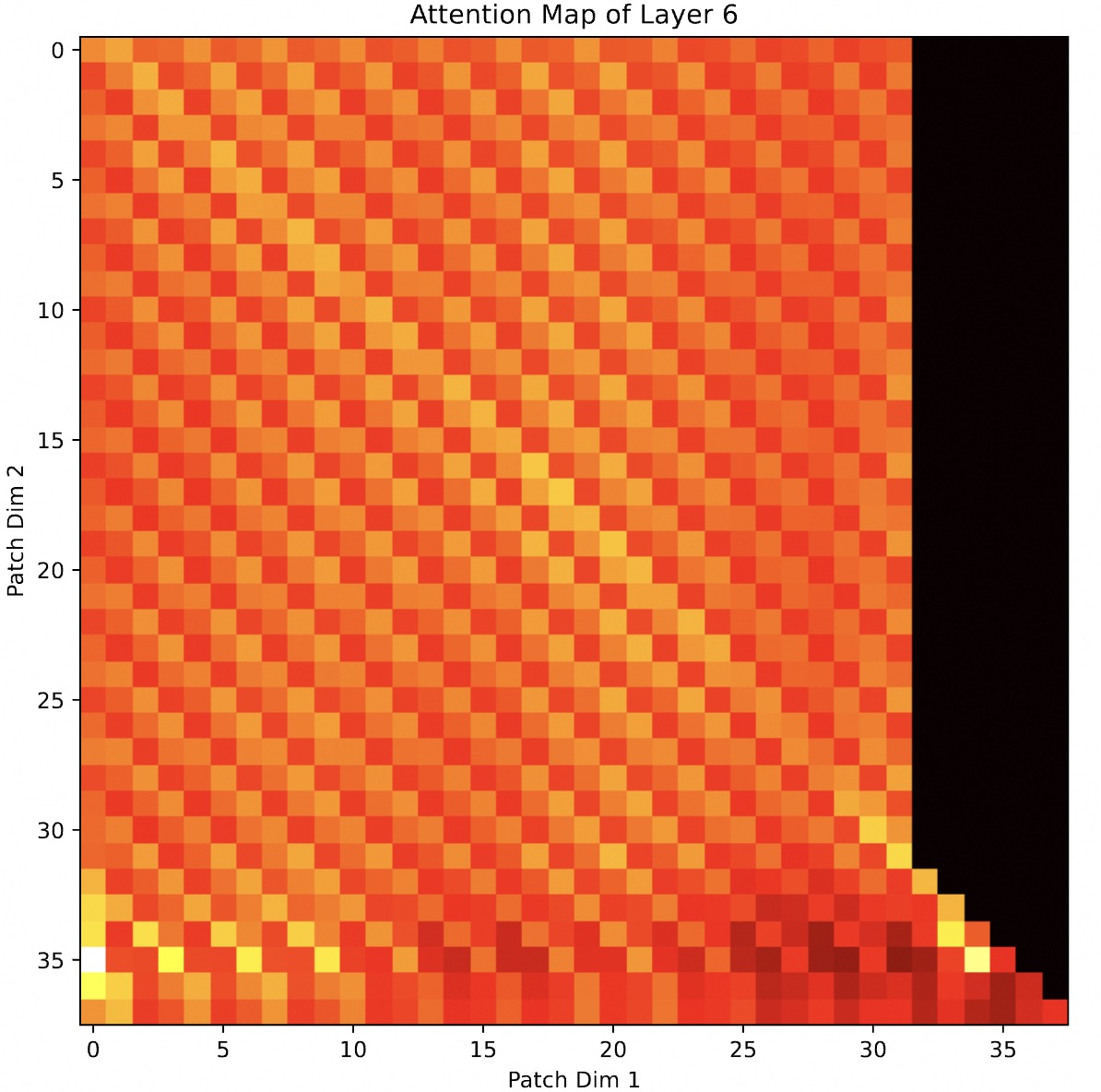}
    }%
    \subfigure[Decoder]{
        \includegraphics[width=0.235\textwidth]{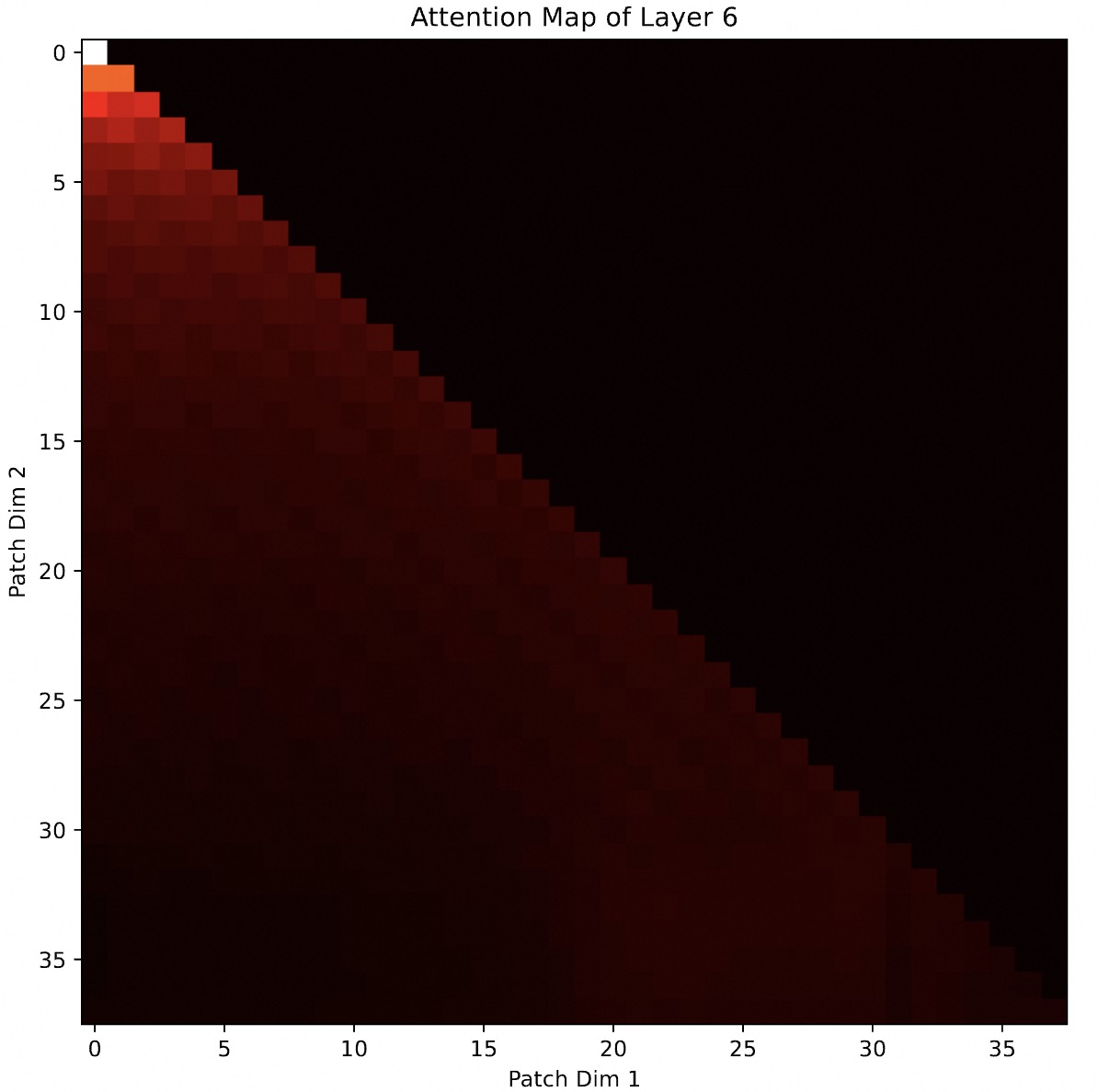}
    }%
    \vspace{-0.7em}
    \caption{Attention maps of the encoder, prefix decoder, and decoder models.}
    \label{fig:Attention_maps}
    \vspace{-1.2em}
\end{figure*}

 Additionally, to further validate the conclusion on the attention mechanisms, we draw the attention maps of the last layer from the encoder, prefix decoder, and decoder models trained on the ETTh1 dataset, as presented in Figure \ref{fig:Attention_maps}. 
\par From the figure, we find that the encoder's attention map presents obvious regularity, which proves that it can well capture patterns and temporal dependencies in time series. In contrast, the decoder model does not show an obvious pattern, which indicates that it is inferior for obtaining temporal dependencies.

\end{sloppypar} \end{document}